\documentclass[twoside]{article}

\usepackage[accepted]{aistats2022}

\usepackage[utf8]{inputenc} 
\usepackage{hyperref}       
\usepackage{url}            
\usepackage{booktabs}       
\usepackage{amsfonts}       
\usepackage{nicefrac}       
\usepackage{microtype}      
\usepackage{multirow}
\usepackage{nicematrix}
\usepackage{tipa}
\usepackage{amsfonts}       
\usepackage{amssymb}
\usepackage{algorithmic}
\usepackage{algorithm}
\usepackage{amsthm}
\usepackage{parallel}
\usepackage{authblk}
\usepackage{blindtext}
\usepackage{wrapfig}
\usepackage{graphicx, color}
\usepackage{amsmath}
\usepackage{verbatim}
\usepackage{mathtools}
\usepackage{hyperref}
\usepackage{subcaption}
\usepackage[english]{babel}
\usepackage{amsthm}
\usepackage[subtle]{savetrees}
\usepackage{float}

\DeclareMathOperator*{\argmin}{arg\,min}
\newtheorem{theorem}{Theorem}[section]
\newtheorem{corollary}{Corollary}[theorem]
\newtheorem{lemma}[theorem]{Lemma}
\newtheorem{proposition}[theorem]{Proposition}
\newtheorem{definition}[theorem]{Definition} 
\newtheorem{remark}[theorem]{Remark} 

\usepackage[round]{natbib}
\setcitestyle{authoryear, open={(},close={)}}

\begin{document}

%

%

\runningtitle{MLDemon}

%
\runningauthor{Ginart, Zhang \& Zou}

\twocolumn[
\aistatstitle{MLDemon: \\
Deployment Monitoring for Machine Learning Systems}
\aistatsauthor{  Antonio A. Ginart$^1$ \And Martin Jinye Zhang$^2$  \And  James Zou$^1$  }

\aistatsaddress{ $^1$Stanford University \And  $^2$Harvard University } ]

\begin{abstract}

 Post-deployment monitoring of ML systems is critical for ensuring reliability, especially as new user inputs can differ from the training distribution.  Here we propose a novel approach, \textsc{MLDemon}, for \textsc{ML} \textsc{De}ployment \textsc{mon}itoring. \textsc{MLDemon} integrates both unlabeled data and a small amount of on-demand labels to produce a real-time estimate of the ML model’s current performance on a given data stream.   Subject to budget constraints, \textsc{MLDemon} decides when to acquire additional, potentially costly, expert supervised labels to verify the model. On temporal datasets with diverse distribution drifts and models, \textsc{MLDemon} outperforms existing approaches. Moreover, we provide theoretical analysis to show that \textsc{MLDemon} is minimax rate optimal for a broad class of distribution drifts.
\end{abstract}

\vspace{-15pt}
\section{INTRODUCTION}
\vspace{-5pt}

When ground-truth labels are not readily available at deployment time, which is often the case if labels are expensive, the most common solution is to use an unsupervised, feature-based anomaly detector \citep{lu2018learning,rabanser2018failing}. In some cases, these detectors work well. However, they may also fail catastrophically since it is possible for model accuracy to fall precipitously without possible detection in just the features. This can happen in one of two ways. First, for high-dimensional data, feature detectors may simply lack a sufficient number of samples to detect all covariate drifts. Second, it is possible that drift only occurs in the conditional distribution of the label $y$ given the features $x$ (this can not be detected without supervision). One potential approach, proposed in \cite{yu2018request}, applies statistical tests to estimate a change in distribution in the features and then requests expert labels only when such a change is detected. While it is natural to assume that a distribution drift in features should be indicative of a drift in the model's accuracy, in reality feature drift is neither necessary\footnote{Drift in the conditional of $y$ given $x$ cannot necessarily be detected from drifts in $x$.} nor sufficient\footnote{Even if drift is present, the model may still generalize well.} as a predictor of accuracy drift In fact, we find that unsupervised anomaly detectors are often brittle. Thus, any monitoring policy that only triggers supervision from feature-based anomaly can fail both silently and catastrophically.

\begin{figure}
\centering
  \includegraphics[width=0.33\textwidth,height=4.5cm]{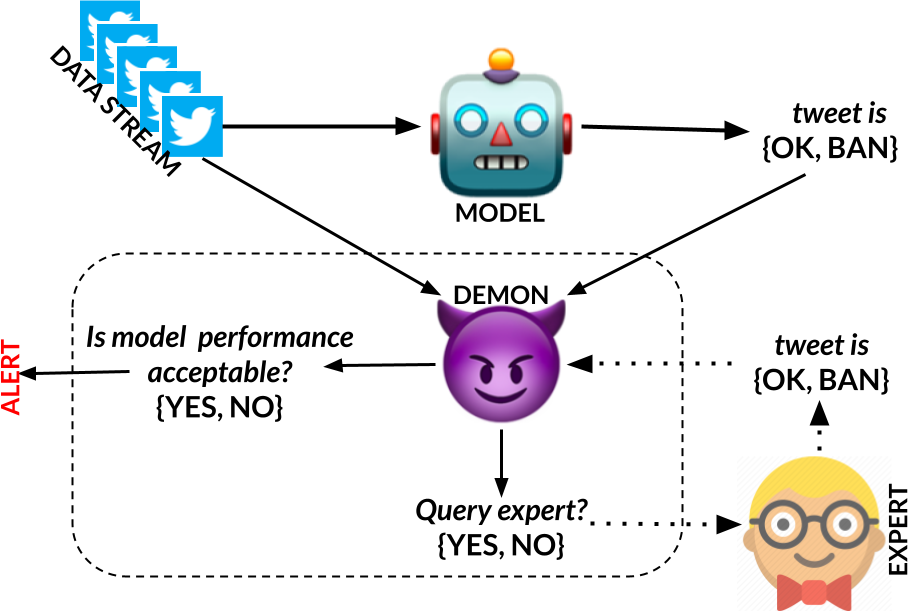}
  \caption{  A schematic for the deployment monitoring workflow. For example, an ML system could be deployed to help automate content moderation on a social media platform. In real-time, a trained \textsc{model} determines if a post or tweet  should result in a ban. A human \textsc{expert} content moderator can review if the content has been correctly classified by the \textsc{model}, though this review is expensive. A \emph{de}ployment \emph{mon}itoring (\textsc{demon}) policy prioritizes expert attention by determining when tweets get forwarded to the \textsc{expert} for labeling. The \textsc{demon} policy also estimates the \textsc{model} performance during deployment which can be used to alert stakeholders if the model performance is not acceptable.}
  \label{fig:problem_schematic}
  \vspace{-15pt}
\end{figure}

Deployment monitoring is a vast topic. In this work, we focus on a particular streaming setting where an automated deployment monitoring policy can query experts for labels during deployment (Fig. \ref{fig:problem_schematic}). The goal of the policy is to estimate the model's real-time accuracy throughout deployment while querying the fewest amount of expert labels. Of course, these two objectives are in contention. We seek to design a policy that can effectively prioritize expert attention at key moments. We focus strictly on monitoring, and do not consider policies that automatically update or debug the deployed model.   

\vspace{-10pt}
\paragraph{Contributions} 
Our contributions are three-fold.

\textbf{(1)} We provide a new mathematical formulation of the ML deployment monitoring problem which is tractable and captures the key trade-off between monitoring cost and risk.

\textbf{(2)}  We theoretically prove that our proposed adaptive monitoring policy, \textsc{MLDemon}, is minimax rate optimal up to logarithmic factors and is superior to prior techniques. 

    
\textbf{(3)} Our experiments reveal that feature-based anomaly detectors can be brittle with respect to real distribution shifts and that \textsc{MLDemon} simultaneously provides robustness to errant detectors while reaping the benefits of informative detectors.

\section{PROBLEM FORMULATION}
We consider a novel online streaming setting \citep{munro1980selection,karp1992line}, where for each time point $t=1,2,\cdots,T$, the data point $X_t \in \mathcal{X}$ and the corresponding label $Y_t \in \{0, 1\}$ are generated from a distribution that may vary over time: $(X_t,Y_t) \sim P_t$. For a given model $f: \mathcal{X} \rightarrow \{0,1\}$, let $\mu_t = \mathbf{Pr}[f(X_t) = Y_t]$ denote its accuracy at time $t$. The total time $T$ can be understood as the life-cycle of the model as measured by the number of user queries. In addition, we assume that we have an anomaly detector, which can depend on both present and past observations and is potentially informative of the accuracy $\mu_t$. For example, the detector can quantify the distributional shift of the feature stream $\{X_t\}$ where a large drift may imply a deterioration of the model accuracy.

We consider scenarios where high-quality labels $\{Y_t\}$ are costly to obtain and are only available upon request from an expert. Therefore, we wish to monitor the model performance $\mu_t$ over time while obtaining a minimum number of labels. We consider two settings that are common in machine learning deployments: 1) point estimation of the model accuracy $\mu_t$ across all time points (estimation problem), 2) determining if the model's current accuracy $\mu_t$ is above or below a user-specified threshold $\rho$ (decision problem). At time $t$, the policy receives a data point $X_t$ and submits a pair of actions $(a_t, \hat{\mu}_t)$, where $a_t \in \{0,1\}$ denotes whether or not to query for an expert label on $X_t$ and $\hat{\mu}_t$ is the estimate of the model's current accuracy. 

It is desirable to balance two types of costs: the average number of queries $Q = \frac{1}{T}\sum_t a_t$ and the monitoring risk. 

In the \textbf{estimation problem} we consider the mean absolute error (MAE) for the monitoring risk:
\begin{align}
    R_{\text{mae}} = \frac{1}{T}\sum_t |\hat{\mu}_t - \mu_t|.
\end{align}
In the \textbf{decision problem}, we consider a binary version $R_{\text{bin}}$ and a continuous version $R_{\text{hinge}}$ for the monitoring risk:
\begin{align}
    R_{\text{bin}} = \frac{1}{T}\sum_t  \mathfrak{R}_t,~~~~ R_{\text{hinge}} = \frac{1}{T}\sum_t \vert \rho - \mu_t\vert \mathfrak{R}_t
\end{align}
where $\mathfrak{R}_t = \mathbf{1}\{\mu_t>\rho,\hat{\mu}_t<\rho \}  + \mathbf{1}\{\mu_t<\rho,\hat{\mu}_t>\rho \}$ is $1$ if the predicted accuracy $\hat{\mu}_t$ and the true accuracy $\mu_t$ incur different decisions when compared to the threshold $\rho$.  
We use $R$ to denote the monitoring risk in general when there is no need to distinguish between the risk functions. Therefore, the combined loss Lagrangian \citep{luenberger1984linear} can be written as:
$
 \mathcal{L} = cQ + R,
$
where $c$ indicates the cost per label query and controls the trade-off between the two types of loss. This Lagrangian implicitly amortizes the query costs and monitoring risks over time. 
Our goal is to design a policy to minimize the expectation of this amortized loss: $\mathbb{E}_P[\mathcal{L}]$.

\vspace{-6pt}
\paragraph{Assumption on Distributional Drift} We are interested in settings for which the distribution $P_t$ varies in time. 
Without any assumption about how $P_t$ changes over time, it is impossible to guarantee that any labeling strategy achieves reasonable performance. Fortunately, many real-world data drifts tend to be more gradual over time. Many ML systems can process hundreds or thousands of user queries per hour, while many real-world data drifts tend to take place over days or weeks. Motivated by this, we consider distribution drifts that are Lipschitz-continuous \citep{o2006metric} over time in total variation\footnote{If $\{ P_t \}$ is $\Delta$-Lipschitz in $d_{\text{TV}}$, then $\{\mu_t \}$ is $\Delta$-Lipschitz in absolute value $|\cdot|$. This gives a natural interpretation to $\Delta$ in terms of controlling the maximal change in model accuracy over time.} \citep{villani2008optimal}: $\mathcal{P} = \{ \{P_t\}_{t=1}^{T} : d_{\text{TV}}(P_t, P_{t-1}) \leq \Delta,  \forall t \}$. The $\Delta$-Lipschitz constraint captures that the distribution shift must happen in a way that is controlled over time, which is a natural assumption for many cases. The magnitude of $\Delta$ captures the inherent difficulty of the monitoring problem; a small $\Delta$ indicates that the deployment is easier to monitor because the stream drifts slowly. For our theory, we focus on asymptotic regret, in terms of $\Delta$, but amortized over the length of the deployment $T$. While our theoretical analysis relies on the $\Delta$-Lipschitz assumption, our algorithm does not require it to work well empirically.  

\paragraph{Feature-Based Anomaly Detection} We assume that our policy has access to a \emph{feature-based anomaly detector} that computes an anomaly signal from the online feature stream. We let $G_t$ denote the \emph{anomaly detection signal} at time $t$. In practice, most anomaly detectors combine a domain-specific summary statistic (or representation) for dimensionality reduction with a statistical metric for testing for drifts in summary statistics (see \cite{rabanser2018failing,yu2018request,wang2019drifted,pinto2019automatic,kulinski2020feature,xuan2020bayesian} for recent examples). Following this general scheme for anomaly detection, we define summary representation $E : \mathcal{X} \rightarrow \mathbb{R}^s$ with dimensionality $s$ that embeds the features in a semantically useful space and (generalized) statistical metric $\mathsf{d}: \mathbb{P}(\mathbb{R}^s ) \times \mathbb{P}(\mathbb{R}^s) \rightarrow \mathbb{R}^{+}$ where $\mathbb{P}(\mathbb{R}^s)$ denotes the set of all distributions over $\mathbb{R}^s$. For any two \emph{detection windows} $S, S' \subset [t]$, we define the detection function $g$:
\begin{align}
    g(S,S') = \mathsf{d}\left(\hat{E}(S), \hat{E}(S')\right)
\end{align}
where $\hat{E}(S)$ denotes the \emph{empirical distribution} of $\{E(X_s):s\in S\}$. At each time $t$, the policy is responsible for providing the detector with the choice of the two detection windows $(S_t, S'_t)$ and the anomaly detection signal returns $G_t = g(S_t,S'_t)$. For example, if $\mathsf{d}(P,Q) = ||\mathbb{E}{P} - \mathbb{E}{Q}||$, then the anomaly signal $G_t$ is equivalent (up to monotonic transformation) to a $Z$-test's $p$-value \citep{montgomery2009engineering} over the detection windows, and if $\mathsf{d}$ is KS-distance \citep{naaman2021tight}, then $G_t$ is equivalent a KS-test's $p$-value \citep{lilliefors1967kolmogorov}. Common summary representations $E$ include embedding layers of deep neural models or even just the model confidence\footnote{ In the case of ML APIs, only the confidence score is typically available \citep{chen2020frugalml}.}.


\paragraph{Linear Detection Condition} The core idea of \textsc{MLDemon} is to combine the information from the expert labels and the detection signal based on the feature stream. Practically speaking, any policy that outright ignores the detector is clearly wasting that side information. On the other hand, any policy that blindly trusts the detector is not robust to model assumption violations. \textsc{MLDemon} strikes a balance between these two extreme positions. There is no universal answer as to how this balance should be struck, as it clearly depends on the particular nature of the detector, drift and problem instance in question. However, it is reasonable to assume that for a well-designed detector (i.e., appropriate choices of $E$ and $\mathsf{d}$), for at least a reasonable fraction of instances, the anomaly signal should roughly correlate with the accuracy drift: $| \mu_t - \mu_{t'}| \approx w \mathsf{d}(E(X_t), E(X_{t'}))$ for some (unknown) constant $w$. The motivating intuition is that the anomaly signal often correlates with the (absolute) accuracy drift. When this is true, one should expect to be able to fit a linear model to partially capture the relationship even if the true relationship is not exactly linear. We do not require the linear model to always hold; given enough evidence against it, MLDemon learns to discard the assumption. We more formally encode this intuition as the \emph{linear detection condition}:

\begin{definition} (Linear Detection Condition)
\label{def:linear_detection_condition}
The linear detection condition holds if with probability at least $q$, for all $S,S'$:
\vspace{-5pt}
\begin{align}
| \mu(S) - \mu(S')| =  w g(S,S') + \mathcal{N}
\label{eqn:linaer_detection_condition}
\end{align} where $\mu(S) = \frac{1}{|S|}\sum_{s \in S} \mu_s$, $\mathcal{N}$ is an independent and identically distributed zero-mean Gaussian\footnote{The Gaussianity of the noise can be replaced with any kind of bounded variance zero mean noise. This would require a more complicated analysis using Berry–Esseen type inequalities but would not change the theoretical asymptotics.} perturbation of arbitrary variance, and $w$ is an arbitrary constant.

\end{definition}
This condition is used for theoretical insights of the proposed algorithm rather than a strictly required assumption. That the condition holds probabilistically is a statement about the assumed distribution over problem instances (and in particular about the relationship between $\{ P_t\}$ and $\{G_t\}$ ) that the policy will encounter.

\section{Algorithms}

We present \textsc{MLDemon} along with two baselines. The first baseline, \textsc{Periodic Querying} (PQ), is a simple non-adaptive policy that periodically queries labels in batches according to a predetermined schedule. The second baseline, \textsc{Request-and-Reverify} (RR), proposed in \cite{yu2018request}, is the previous state-of-art to our problem (code sketch in Alg. \ref{alg:RR}). All of the policies run in constant space and amortized constant time --- an important requirement for a scalable long-term monitoring system. 

\vspace{-5pt}
\paragraph{\textsc{Periodic Querying}}  works for both the estimation problem and the decision problem. As shown in Alg. \ref{alg:PQ}, given a budget $B$ for the average number of queries per round, PQ periodically queries for a batch of $n$ labels in every $\frac{n}{B}$ rounds, and uses the estimate from the current batch of labels for the entire period. \footnote{Another possible variant of this policy queries once every $\frac{1}{B}$ rounds and combines the previous $n$ labels. When $\Delta$ is known or upper bounded, we may instead set the query rate to guarantee some worst-case $\epsilon$ monitoring risk.}

\begin{algorithm}[H]
\scriptsize
\textbf{Inputs:} At each time $t$, the previously observed data points and queried labels

\textbf{Outputs:} At each time $t$, $ a_t \in  \{0,1\} $, $ \hat \mu_t \in [0,1]$

\textbf{Hyperparameters:} Window length $n$, budget $B \in [0,1]$ 

\begin{algorithmic}
\STATE \textbf{do:}
\vspace{-5pt}
\begin{enumerate}
    \item Query ($a_t \gets 1$) for $n$ consecutive labels and then do not query ($a_t \gets 0$) for $(1/B -1)n$ rounds \vspace{-5pt}
    \item Compute $\hat{\mu}_t$ from most recent $n$ labels as empirical mean
\end{enumerate}
\STATE \textbf{repeat}

\end{algorithmic}
\caption{\textsc{Periodic Querying}}\label{alg:PQ}
\end{algorithm}


\vspace{-20pt}
\paragraph{\textsc{Request-and-Reverify}} sets a predetermined threshold $\phi \geq 0$ for anomaly signal $G_t$ and queries for a batch of $n$ labels whenever the  threshold is exceeded by the anomaly signal $G_t \geq \phi $. As for the anomaly detector, RR applies a statistical test on a sliding window of model confidence scores. As discussed in Section 2, the choice of test corresponds to a particular statistical distance $\mathsf{d}$. By varying the threshold $\phi$, RR  can vary the number of labels queried in a deployment.\footnote{The natural interpretation is that $\phi$ corresponds to a particular threshold $p$-value for the statistical test that triggers a new label batch.} In optimistic circumstances, RR is the essentially optimal policy. For example, consider the problem instance: $\mu_t = 1$ for $t < \tau$, $\mu_t = 0$ for $t \geq \tau$, and $G_t = 1$ at $t = \tau$, $G_t = 0$ otherwise. On the other hand, while training data can be used to calibrate the threshold, in our theoretical analysis we show that for any $\Delta >0$, RR cannot provide a non-trivial worst-case guarantee for monitoring risk, regardless of the choice of anomaly detector. This worst-case is realized when the detector is errant. 

\begin{algorithm}[H]
\scriptsize
\textbf{Inputs:} At each time $t$, the previously observed data points and queried labels, and anomaly signal $G_t$

\textbf{Outputs:} At each time $t$, $ a_t \in  \{0,1\} $, $ \hat \mu_t \in [0,1]$

\textbf{Hyperparameters:} Window length $n$, threshold $\phi$ 


\begin{algorithmic}
\STATE $G_t \gets \text{Compute anomaly score} $
\IF{$ G_t \geq \phi$ and not currently querying }

    \STATE (1) Query ($a_t \gets 1$) for a batch of $n$ consecutive labels 
    \STATE (2 $\hat{\mu}_t \gets$ empirical mean of  $n$  most recent observed outcomes

\ELSE
\STATE Do not query for labels ($a_t \gets 0$) and keep $\hat \mu_t$ fixed
\ENDIF
\end{algorithmic}
\caption{\textsc{Request-and-Reverify} \label{alg:RR}}
\end{algorithm}

\vspace{-15pt}
\paragraph{Detection Windows} Recall that any policy specifies two detection windows at each time, $S_t, S'_t$.  PQ does not use the anomaly signal, and thus the detection windows are irrelevant. For RR and \textsc{MLDemon}, it is natural in both cases to use the same strategy, defined in Alg. \ref{alg:detection_windows}.

\vspace{-3pt}
\begin{algorithm}[H]
\scriptsize
\textbf{Hyperparameters:} Window length $n$

\textbf{Outputs:} At each time $t$, detection windows $S_t, S'_t \subset [t]$ 

\begin{algorithmic}

\STATE $\tau \gets \text{time since most recent query batch}$
\STATE $S_t \gets \{t -\tau -n,\dots, t-\tau \}$
\STATE $S'_t \gets \{t -n,\dots , t \}$
\STATE \textbf{return } $(S_t, S'_t)$

\end{algorithmic}
 \caption{Detection Windows }\label{alg:detection_windows}
\end{algorithm}
\vspace{-10pt}

\paragraph{\textsc{MLDemon}} follows a periodic query cycle like PQ. However, \textsc{MLDemon} also fits and evaluates a linear estimate of the accuracy drift based on the anomaly signal. If and when \textsc{MLDemon} becomes sufficiently confident in the model, \textsc{MLDemon} will extend the period in between querying if the anomaly signal yields enough evidence that the model accuracy is sufficiently stable. \textsc{MLDemon} does this by independently generating confidence intervals around $\hat{\mu}_t$ based on both the expert labels and on the anomaly signal. \textsc{MLDemon} then combines these independent intervals with Bayes rule \citep{tipping2003bayesian}. We think of Alg. \ref{alg:mldemon} as a routine that runs at each time $t$.






As discussed, $E$ and $\mathsf{d}$ are considered part of the problem instance via the given detector. Thus, one of \textsc{MLDemon}'s jobs is to select detection windows $(S_t,S'_t)$ (see Alg. \ref{alg:detection_windows}). Whenever a new batch of label queries is completed, \textsc{MLDemon} has a good estimate of $\hat{u}_t$, and thus it is a good time to fit the linear detection model using the anomaly signal. Concretely, \textsc{MLDemon} applies an ordinary least squares (OLS) update to $\hat{w}$ based on $(G_t,\partial_t\mu)$ where $\partial_t\mu = |\hat{\mu}(S_t) - \hat{\mu}(S'_t)|$. 

\vspace{-3pt}
\begin{algorithm}[h]
\scriptsize

\textbf{Inputs:} Anomaly signal $\{G_t\}$, point estimate history $\{\hat \mu_t\}$,  

\textbf{Outputs:} At each time $t$,  action $ a_t \in  \{0,1\} $, accuracy estimate $ \hat \mu_t \in [0,1]$

\textbf{Hyperparameters:}  Batch size $n$,  risk tolerance $\epsilon$, query period $\alpha$, drift bound $\Delta$, linear detection prior $q$ 
\begin{algorithmic}

\STATE $N \gets 0$
\STATE \textbf{do:}
\vspace{-5pt}
\begin{enumerate}
    \item \texttt{Query} ($a_t \gets 1$) for $n$ consecutive labels and then \texttt{do not query} ($a_t \gets 0$) for $\alpha \cdot n$ rounds \vspace{-5pt}
    \item compute $\hat{\mu}_t$ from the most recent $n$ labels as empirical mean
\end{enumerate}

\IF{At the \emph{end} of a \texttt{query} period}
\STATE $S \gets \{\text{previous label batch} \}$
\STATE $S' \gets \{\text{current label batch} \}$
\STATE $G_t \gets g(S,S')$
\STATE $\partial_t^\mu \gets |\hat{\mu}(S) - \hat{\mu}(S')|$
\STATE $\hat{w}_t \gets \text{OLS update on } \hat{w}_{t-1} \text{ given new data point } (G_t,\partial_t^\mu)$
\STATE $N \gets N + 1$
\ENDIF

\IF{At the \emph{end} of a \texttt{do-not-query} period}

\STATE $\mathsf{se}_t \gets \text{std. err. of the forecast \citep{buteikis2019practical} using OLS}$
\STATE $\tau \gets \text{time since most recent query}$
\STATE $G_t \gets \text{Compute anomaly score} $
\STATE $\nu_t \gets \Delta\cdot (\tau + (n+1)/2)$
\IF{\text{decision problem}}
\STATE $ \ell_t \gets \max\{ |\hat{\mu}_t - \rho| - \epsilon, 0\}/\Delta$
\ELSE
\STATE $ \ell_t \gets 0$
\ENDIF
\STATE $p_{\text{lbl}} \gets 1 - 2\exp(-2n(\nu_t - \epsilon + \ell_t - n\Delta)^2) $
\STATE $p_{\text{det}} \gets2 -2 \cdot \mathtt{Student\_t\_cdf\_with\_(N-2)\_deg\_of\_freedom}\left(
      \frac{\epsilon + \ell_t -n\Delta}{\mathsf{se}_t} \right)$

\STATE $p_t \gets q\left(\frac{p_{\text{lbl}}p_{\text{det}}}{p_{\text{lbl}}p_{\text{det}} + (1-p_{\text{lbl}})(1-p_{\text{det}})} \right) + (1-q)p_{\text{lbl}} $
\IF{$p_t \geq 1 - \epsilon$ and $\epsilon - \nu_t + \ell_t - n\Delta > 0$}
\STATE Extend the \texttt{do not query} period by 1: 
\STATE $a_t \gets 0$
\ELSE
\STATE End the \texttt{do not query} period and begin to \texttt{query}: 
\STATE $a_t \gets 1$
\ENDIF

\ENDIF

\STATE \textbf{repeat}


\end{algorithmic}
 \caption{\textsc{MLDemon} }\label{alg:mldemon}

\end{algorithm}

\textsc{MLDemon} aggregates the information from recent labels with the anomaly signal from the detector to yield confidence intervals around $\hat{\mu}_t$. The label information, via the queries, is used to construct an $\epsilon$-width interval at confidence level $1 - p_{\text{lbl}}$ (that is what $\ell_t$ and $\nu_t$ are used for). The precise formula used to compute $p_{\text{lbl}}$ is based on our novel extension of Hoeffding's Inequality to $\Delta$-Lipschitz setting (see Appendix). The feature information, via anomaly detection  signal, is used to construct an $\epsilon$-width interval at confidence level $1-p_\text{det}$ (assuming the linear detection condition). The precise formula used to compute $p_{\text{det}}$ is based on the standard prediction interval using a Student-t distribution (which is valid when $\mathcal{N}$ is Gaussian). These two intervals are independent and can be combined with the Bayes rule using the prior $q$ to obtain the joint confidence level $p_t$.

 The anomaly signal may not correlate with changes in the model's accuracy, so we would also like to incorporate some robustness to counteract the possible failure event of an uninformative or errant detector. Therefore, \textsc{MLDemon} treats $q$ as a hyperparameter that encodes the system's prior belief in the informativeness of the detection signal. Even if the $q$ given to \textsc{MLDemon} is incorrect, \textsc{MLDemon} will still behave reasonably. Unlike RR, \textsc{MLDemon} never assumes the detector is trustworthy, even if $q\gets1$. \textsc{MLDemon} only trusts the detector it verifies that the detector predicts the drift in $\hat{\mu}$.  

For \textsc{MLDemon}, the key variables are (1) monitoring risk tolerance $\epsilon$ that is a desired upper bound on $\mathbb{E}[R]$, (2) the window length for batches of label queries, (3) the query rate, given by $\Theta(1/\alpha)$ for $\alpha \geq 1$, and (4) a Lipschitz constant on the drift, $\Delta$. To deploy \textsc{MLDemon} one must specify two out of $(\epsilon, \Delta, n, \alpha)$.\footnote{With the exception of the choice of pair $\epsilon$ and $n$, which does not allow \textsc{MLDemon} to pin down $\Delta$ and $\alpha$.} One could also specify $q$ as the prior for the linear detection condition, but, as we shall see, $q \gets 1- \epsilon$ is also a solid choice theoretically and empirically. Upon specifying these two variables as hyperparameters, \textsc{MLDemon} can automatically solve for the remaining two (see Appendix for more details concerning hyperparameter selection).

 For the decision problem, we can additionally increase the query period based off the estimated margin to the target threshold using our estimate of $|\hat \mu_t - \rho|$. With a larger margin, we need a looser confidence interval to guarantee the same monitoring risk. This translates into fewer label queries while still preserving a risk tolerance upper bound of $\epsilon$.

\section{Theoretical Analysis}


\paragraph{Minimax Analysis} Our asymptotic analysis in this section is concerned with an asymptotic rate in terms of small $\Delta$  and amortized by a large $T$. When using asymptotic notation, by loss $\mathcal{L} = O(\Delta^k)$ we mean $\lim_{\Delta \rightarrow 0} \lim_{T \rightarrow \infty} \mathcal{L} \leq c_0 \Delta^k$, for some constant $c_0>0$. Recall that amortization is implicit in the definition of $\mathcal{L} = cQ + R$ (defined in Section 2). We use tilde notation (for example, $\widetilde{O}$) to denote the omission of logarithmic factors. We let $\mathcal{L}_g^\pi$ be the combined loss when using policy $\pi$ and anomaly detector $g$. Recall that a detector $g$ is defined by representation $E$ and statistical distance $\mathsf{d}$. When using MAE risk, we only consider estimation problems, and when using hinge risk, we only consider decision problems. All the proofs are in the appendix. 


\begin{theorem} 
\label{thm:worst_case_regret_main}
Let $\mathcal{P} = \mathbf{Lip}(\Delta)$ be the set of $\Delta$-Lipschitz drifts and let $\Pi$ be the space of deployment monitoring policies. For both MAE risk and hinge risk, for any model $f$ and anomaly detector $g$:

(i) PQ has a worst-case expected loss $\sup_{P \in \mathcal{P}} \mathbb{E}_{P}[\mathcal{L}^{\mathbf{PQ}}_{g}] = \widetilde{O}\big( \Delta^{1/4}\big)$ ;

(ii) When $0 \leq q < 1$ is constant and $\epsilon = \Theta(\Delta^{1/4})$,  \textsc{MLDemon} has a worst-case expected loss $\sup_{P \in \mathcal{P}} \mathbb{E}_{P}[\mathcal{L}^{\mathbf{MLD}}_{g}] = \widetilde{O}\big( \Delta^{1/4}\big)$;

\normalsize
(iii) RR has a worst-case expected loss $\sup_{P \in \mathcal{P}} \mathbb{E}_{P}[\mathcal{L}^{\mathbf{RR}}_{g}] = \Theta(1)$;

(iv) No policy can achieve a better worse-case expected loss than \textsc{MLDemon} and PQ: $\inf_{\pi} \text{ } \sup_{P} \mathbb{E}_{P}[\mathcal{L}^{\pi}_{g}] = \Omega\left( \Delta^{1/4}\right)$.
\end{theorem}

\textsc{MLDemon} is minimax rate optimal up to logarithmic factors. This analysis treats  \textsc{MLDemon}'s $q$ as a non-asymptotic hyperparameter, but Thm. \ref{thm:worst_case_regret_main} does \emph{not} assume the linear detection condition. In contrast to the robustness of \textsc{MLDemon}, RR can fail catastrophically with any detector. For hard problem instances, the anomaly signal is errant and the threshold margin is small, so \textsc{MLDemon} cannot outperform PQ from a minimax perspective.

\begin{lemma}
\label{lemma:worst_case_tradeoff}
Under the conditions of Thm. \ref{thm:worst_case_regret_main}, for both MAE and hinge risk, PQ and \textsc{MLDemon} achieve a worst-case expected monitoring risk of $O(\epsilon)$ with a query rate of $O(\frac{\Delta \log(1/\epsilon)}{\epsilon^3})$ and no policy can achieve a query rate of $\omega(\Delta/\epsilon^3)$ with monitoring risk $O(\epsilon)$.
\end{lemma}

Lem. \ref{lemma:worst_case_tradeoff} is used to prove Thm. \ref{thm:worst_case_regret_main}, but we include it here because it is of independent interest to understand the trade-off between monitoring risk and query costs and it also gives intuition for Thm. \ref{thm:worst_case_regret_main}. The emergence of the $\Delta^{1/4}$ rate also follows from Lem. \ref{lemma:worst_case_tradeoff} by considering the combined loss $\mathcal{L}$ optimizing over $\epsilon$ to minimize $\mathcal{L}$ subject to the constraints imposed by Lem. \ref{lemma:worst_case_tradeoff}.\footnote{There remains a gap of order $\log(1/\Delta^{1/4})$ in between our achievable expected loss and our lower bound. It would be interesting to close this gap.} Lem. \ref{lemma:worst_case_tradeoff} itself follows from an analysis that pairs a lower bound derived with Le Cam's method \citep{yu1997assouad} and an upper bound constructed with an extended Hoeffding's inequality \citep{hoeffding1994probability} that handles samples with $\Delta$-Lipschitz drift.

\vspace{-4pt}
\paragraph{Optimistic Analysis} We investigate how much \textsc{MLDemon} can improve over PQ by using the linear detection condition. When it holds with probability $q$, \textsc{MLDemon} can guarantees a worst-case monitoring risk $\epsilon$ while reducing the labeling rate significantly if: (i) the noise magnitude of $\mathcal{N}$ is small and (ii) the accuracy drift $\{\mu_t\}$ is stable. Condition (i) enables \textsc{MLDemon} to fit the linear model with low forecast error and condition (ii) enables \textsc{MLDemon} to extend the query period since the policy can confidently forecast that $\{\mu_t\}$ is not drifting. For the following theorem, it will be useful to recall that $Q^{\mathbf{PQ}} = \sup_{P \in \mathcal{P}} \mathcal{Q}^{\mathbf{MLD}}_{g}$ since \textsc{MLDemon} only ever \emph{reduces} the query rate compared to PQ.

\begin{theorem} 
\label{thm:linear_case_regret_main}
Under the conditions of Thm. \ref{thm:worst_case_regret_main}, the following hold if, additionally,  $(E,\mathsf{d})$ satisfies the linear detection condition with probability $q$:

(i) if using MAE risk and $q < 1 - \epsilon $ then, $$  \frac{3}{13} - \widetilde{O}(\Delta^{1/4}) \leq \inf_{P \in \mathcal{P}} \mathbb{E}_P \left[\frac{\mathcal{Q}^{\mathbf{MLD}}_{g} }{\mathcal{Q}^{\mathbf{PQ}}_{g}}\right]\leq 
\frac{9}{19} -   \widetilde{O}(\Delta^{1/4})$$

(ii) if using hinge risk and $q < 1 - \epsilon $ then,

$$ \inf_{P \in \mathcal{P}} \mathbb{E}_P \left[\frac{\mathcal{Q}_{g}^{\mathbf{MLD}} }{\mathcal{Q}_{g}^{\mathbf{PQ}}}\right] = \widetilde{O}(\Delta^{1/4})$$

(iii) if  $q \geq 1 - \epsilon$, then there exist problem instances $P$ for which $Q_{g}^{\mathbf{MLD}}/Q_{g}^{\mathbf{PQ}} = 0$ almost surely.

\end{theorem}

Thus, there are two regimes for \textsc{MLDemon}. When $q \rightarrow 1$ (part iii), there exist problem instances for which \textsc{MLDemon} exhibits behavior like RR, namely, the possibility of never querying for a label after a certain point in the stream (which is a dangerous behavior). When $q$ is constant (parts i \& ii), \textsc{MLDemon} is still minimax rate optimal as stated in Thm. \ref{thm:worst_case_regret_main}, but can reduce the number of label queries by up to a factor of about 2-4$\times$ when using MAE risk and $\Delta$ is small and even more so when using hinge risk.

\paragraph{Average-Case Analysis} For the hinge risk, there is gap a in asymptotic query rates for minimax and optimistic situations. To investigate this further, we perform an average-case analysis with a stochastic model implying a distribution over problem instances in $\mathcal{P}$. Our model assumes the following law, denoted as $\mathcal{S}$, for generating the sequence $\{\mu_t\}$ from any arbitrary initial condition $\mu_0$: $\mu_t = \mathbf{min}\{ \mathbf{max} \{ \mu_{t-1} + \mathbf{Unif}(-\Delta,\Delta),1\},0\}$. The accuracy drift is modeled as a simple random walk. As discussed in \cite{szpankowski2011average} the maximum entropy principle (used by our model at each time step under the $\Delta$-Lipschitz constraint) is often a reasonable stochastic model for average-case analysis.  



 \begin{theorem} 
 \label{thm:avg_case_main} For hinge risk and  model $f$, and detector $g$, when $0 \leq q < 1$ is constant and $\epsilon = \Theta(\Delta^{1/4})$: 
 
 $$ \frac{\mathbb{E}_{P \sim \mathcal{S}} \mathcal{L}_g^{\mathbf{MLD}}}{ \mathbb{E}_{P \sim \mathcal{S}}  \mathcal{L}_g^{\mathbf{PQ}}} \leq \widetilde{O}(\Delta^{1/12})$$

\end{theorem}
The reason we have a better asymptotic gain in the decision problem is illuminated below in Lem. \ref{lemma:avg_q_decision}.
 \begin{lemma}
 \label{lemma:avg_q_decision}
 For hinge risk under model $\mathcal{S}$, \textsc{MLDemon} achieves an expected monitoring  hinge risk $O(\epsilon)$ with an amortized query amount  $\widetilde{O}\left(\Delta/\epsilon^2\right)$. 
 \end{lemma}

\textsc{MLDemon} can save an average  $1/\epsilon$ factor in query cost, which translates into the rate improvement in Thm. \ref{thm:avg_case_main}. \textsc{MLDemon} does this by leveraging the margin between estimate $\hat \mu$ and threshold $\rho$ to increase the confidence interval width around the estimate without increasing risk. Note that when using the minimax optimal hyperparameters $\epsilon = \Theta(\Delta^{1/4})$, then Lem. \ref{lemma:avg_q_decision} also implies that the expected ratio of query rates behaves like the optimistic ratio: $\mathbb{E}_{\mathcal{S}}[{Q^{\mathbf{MLD}}/Q^{\mathbf{PQ}}}] = \mathbb{E}_\mathcal{S}[Q^{\mathbf{MLD}}]/Q^{\mathbf{PQ}} = \widetilde{O}(\Delta^{1/4})$ (see appendix for more details).



 \section{Experiments}
 Our asymptotic analysis indicates that \textsc{MLDemon} can achieve robustness close to that of PQ yet also reap the benefits of an informative detector like RR. In this section, we implement the algorithms on real data streams as a proof-of-concept.

 \paragraph{Data Streams}  We benchmark the 3 policies on 3 data stream benchmarks (summarized below and in Table \ref{table:data_stream_overview}).
\vspace{-1pt}



\textbf{(1)} \texttt{SPAM-CORPUS} \citep{katakis2006dynamic}: A non-stationary data set for detecting spam mail over time based on text. It represents a real, chronologically ordered email inbox from the early 2000s and is a canonical benchmark for non-stationary learning.
     
  \textbf{(2)} \texttt{WEATHER-AUS}\footnote{https://www.kaggle.com/jsphyg/weather-dataset-rattle-package}:  A  non-stationary data set for predicting if it will rain in Australia based on other weather and geographic features. The data is gathered from a range of locations and time spanning years.
      
  \textbf{(3)} \texttt{FACE-R} \citep{wang2020masked}: A data set that contains multiple images of hundreds of individuals, both masked and unmasked. The distribution drift mimics the onset of a pandemic by increasing the percentage of masked individuals. Initially, the masked percentage grows slowly, followed by a sharp increase.
 
 \begin{table*}[h]
\centering
\footnotesize 
\begin{tabular}{ |p{2.3cm}||p{0.9cm}|p{1.2cm}|p{2.2cm}|p{1.5cm}|p{1.5cm}||p{2.0cm}| }
 \hline
 \multicolumn{7}{|c|}{Data Stream Details} \\
 \hline
 Data Stream & $T$ & \# class &  Model & $\mu_0$ & $\mathsf{se}$ & $(\varepsilon_{\min} , \varepsilon_{\max})$ \\
 \hline

\texttt{SPAM-CORPUS} & 7,300 & 2 & Logistic Reg. &  $92\%$ & 0.0695 & $(0.0674,
0.134)$\\ 
\texttt{WEATHER-AUS} & 45,000 & 2 & Logistic Reg. &  $86\%$  & 0.175  & $(0.091, 0.134)$\\ 
\texttt{FACE-RECOG} & 40,000 & 400 & Residual CNN & $38\%$ &0.0304 & $(0.053,0.097)$\\ 

 \hline
\end{tabular}
  \caption{ Three data streams used in our empirical study. $T$ is the length of the stream in our benchmark. The initial test accuracy on the training distribution at time $t=0$ is denoted by $\mu_0$ and standard error of the forecast by an OLS linear model following Eqn. \ref{eqn:linaer_detection_condition} is denoted by $\mathsf{se}$. The minimal and maximal empirical average monitoring risk are denoted by $\varepsilon_{\min}$ and $\varepsilon_{\max}$, respectively. Also reported are the number of classes in the classification task and the classifier used.}
    \label{table:data_stream_overview}
    \vspace{-8pt}
\end{table*}

\vspace{-7pt}
\paragraph{Anomaly Detector}  Following \cite{yu2018request,rabanser2018failing}, for \texttt{SPAM-CORPUS} and \texttt{WEATHER-AUS}, we take let representation $E(X)$ be the model's confidence on a particular data point $X$, as given by the model logits. This is a popular choice and is often the only possible choice when using inference APIs. For \texttt{FACE-RECOG}, we use the face embedding $E(X)$ produced by a residual CNN for face $X$. For the choice of $\mathsf{d}$, we follow \cite{rabanser2018failing} and simply take $\mathsf{d}(P,Q) = ||\mathbb{E}{P} - \mathbb{E}{Q}||$ (equivalent to a $Z$-test on the empirical means).

The informativeness of the anomaly detection signal is a key aspect of a problem instance. Over the entire stream, we can globally quantify this with the \emph{standard error of the forecast} \citep{buteikis2019practical} (herein denoted $\mathsf{se}$) of an OLS linear fit of the anomaly signal. Smaller $\mathsf{se}$ indicate more accurate predictions from the linear model. While $\mathsf{se}$ is a reasonable global metric, it does not capture all of the nuances. Notice that the realized informativeness of the anomaly signal can depend on the label query rate. Taking RR as an example, suppose $\phi$ is high enough such that it is exceeded but once and RR only queries for one batch of labels. Further suppose that this occurs at the precisely optimal moment given that the policy will only query one batch of labels. For this example, it would seem the detector was highly informative. However, in the same example instance, it could be that even slightly lowering $\phi$ causes RR to make many additional label batch queries at particularly sub-optimal moments (for example, if the stream is not drifting at those moments). At a higher label rate, the same detector may seem mostly uninformative or even outright errant.


\vspace{-5pt}
\paragraph{Training, Implementation \& Hyperparameters} For the logistic regression models, we train models on the first $5\%$ of the drift, then treat the rest as the deployment test. For $\texttt{FACE-R}$, we use an open-source facial recognition package\footnote{https://github.com/ageitgey/face\_recognition} that is powered by a pre-trained residual CNN \citep{he2015deep} that computes face embeddings.

Each dataset is a time-series of labeled data $\{(X_t,Y_t)\}$. For each dataset, we generate 8 streams by randomly shuffling the data locally. As a proxy for the true $\{\mu_t\}$, which is unknown for real data, we use compute a moving average for the empirical $\mu_t = \frac{1}{\ell}\sum_{i=0}^{\ell} \mathbf{1}\{f(X_{t-i}) = Y_{t-i}\}$ with sliding window length $\ell = 250$. For all methods, we fix the label query batch size $n = 35$. This was a neutral choice made by tuning for the best batch size for PQ on a synthetic dataset. For PQ, we sweep the amortized query budget $B$. For \textsc{MLDemon}, with $n$ fixed, for consistency with PQ, we choose to sweep query rate parameter $\alpha$.  For RR we also sweep the anomaly threshold $\phi$. See Appendix for more details.

\begin{figure*}[h]
     \centering
     \begin{subfigure}[b]{0.49\textwidth}
         \centering
         \includegraphics[width=\textwidth]{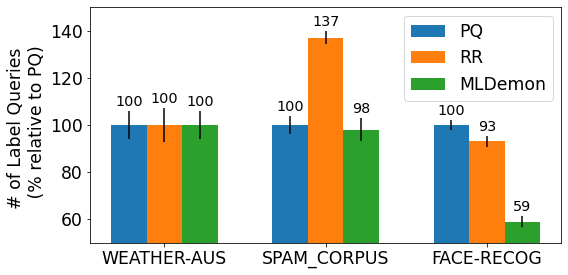}
         \caption{ Lower monitoring risk: $\eta= 0.15 $. The baseline avg. query rate for PQ is $7.4\%$ on \texttt{WEATHER-AUS}, $7.1\%$ on \texttt{SPAM-CORPUS}, and $1.3\%$ on \texttt{FACE-RECOG}. }
         \label{fig:ppl_eps_dpsgd}
     \end{subfigure}
     \hfill
     \begin{subfigure}[b]{0.49\textwidth}
         \centering
         \includegraphics[width=\textwidth]{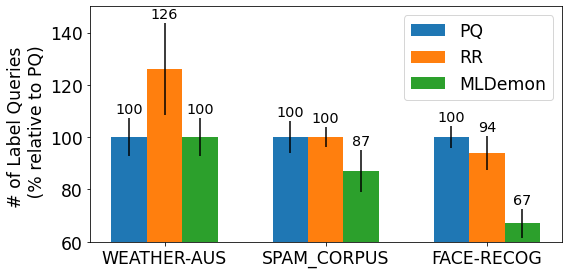}
         \caption{Moderate monitoring risk: $\eta= 0.30$. The baseline avg. query rate for PQ is $4.1\%$ on \texttt{WEATHER-AUS}, $6.3\%$ on \texttt{SPAM-CORPUS}, and $0.5\%$ on \texttt{FACE-RECOG}.}
         \label{fig:ppl_eps_sa}
     \end{subfigure}

\label{fig:three graphs}
\caption{ \small Number of label queries needed by PQ, RR, and \textsc{MLDemon} to achieve a target monitoring risk. Each target monitoring risk (MAE) is computed by $\varepsilon = \eta(\varepsilon_{\max} - \varepsilon_{\min}) + \varepsilon_{\min}$ for each benchmark. Error bars are interpolated std. err. of the mean.}
\vspace{-5pt}
\end{figure*}

\vspace{-4pt}

\vspace{-3pt}
\paragraph{Minimal and Maximal Monitoring Risk} We define the \emph{maximal monitoring risk} as $\varepsilon_{\max} = \frac{1}{T}\sum_{_t}|\mu_0 - \mu_t|$. The maximal monitoring risk serves as a good upper bound for the monitoring risk $R$ because it is the risk of a policy that never queries for a single label and simply relies on the initial test accuracy. Similarly, for a given batch size  $n$ (in this case $n = 35$), the \emph{minimal monitoring risk} is defined as $\varepsilon_{\min} = \frac{1}{T}\sum_{_t}|\frac{1}{\tau}\sum_{\tau=t-n}^t Y_\tau - \mu_t|$. The minimal monitoring risks serves as a good lower bound for $R$ because it the risk of a policy that queries for every single label while using a given batch size for estimating $\hat{\mu}$.

\vspace{-3pt}
\paragraph{Results (Fig. 2)} \textsc{MLDemon} performs at least as well as PQ and RR, and in some cases, performs significantly better (around a third fewer labels than RR and up to a $40\%$ reduction in labels compared to PQ). When the anomaly detector is errant (as evidenced by RR performing significantly worse than PQ), we \textsc{MLDemon} behaves nearly identically to PQ, because the OLS linear fit is not dependable. If RR performs equivalently to PQ, \textsc{MLDemon} may sometimes also perform equivalently, but might also moderately improve upon both.

On \texttt{WEATHER-AUS}, the $\mathsf{se}$ is not low enough for \textsc{MLDemon} to extend the query period, and thus it behaves like PQ. For low monitoring risk $(\eta = 0.15)$, RR behaves comparably, although not identically. At $\eta = 0.3$, RR performs poorly. On \texttt{SPAM-CORPUS}, \textsc{MLDemon} is able slightly outperform the baselines at $\eta = 0.15$ and more substantially at $\eta = 0.3$ (recall that detector quality depends on the query rate and thus the monitoring risk). On \texttt{FACE-RECOG}, the face embeddings tend to have an easily detectable signal as to whether or not the individual is masked, which results in a larger improvement for RR and \textsc{MLDemon}. However, RR fails to detect gradual decreases in accuracy early on in the stream, only querying near the end when the larger transition occurs. \textsc{MLDemon}'s surveillance querying captures the slow drifts that might be too gradual for the anomaly signal.


The experimental findings support the theoretical conclusion that it is a risky policy to purely rely on potentially brittle anomaly detection instead of balancing surveillance queries with anomaly-driven queries. Of course, we caution that the quality of the anomaly detector is not the only factor in a policy's performance. The stability of the drift also plays a major role.

\vspace{-2pt}
\section{Discussion}
\vspace{-2pt}


\paragraph{Related Works} While our problem setting is novel, there are a variety of settings relating to ML deployment and distribution drift. One such line of work focuses on reweighting data to detect and counteract label drift \citep{lipton2018detecting,garg2020unified}.
Another related problem is when one wants to \emph{combine} expert and model labels to maximize the accuracy of a joint classification system \citep{cesa1997use,farias2006combining, morris1977combining, fern2003online}. The problem is similar in that some policy needs to decide which user queries are answered by an expert versus an AI, but the problem is different in that it is interesting even in the absence of online drift or high labeling costs. It would be interesting to augment our formulation with a reward for the policy when it can use an expert label to correct a user query that the model got wrong. This setting would combine both the online monitoring aspects of our setting along with ensemble learning \citep{polikar2012ensemble,minku2009impact} under concept drift with varying costs for using each classifier depending on if it is an expert or an AI.  

 Complementary to \textsc{MLDemon}'s focus on the supervision-monitoring trade-off is work by  \cite{shao2020increasing} and \cite{pinto2019automatic}, which are more focused on the alerting, explanation, and drift detection aspects of the problem without investigating the supervision-monitoring trade-off. Another related approach, explored in \cite{schelter2020learning}, is to try to corrupt training data synthetically in order to fit a predictor that generalizes to natural drift in deployment.

Adaptive sampling rates are a well-studied topic in the signal processing literature \citep{dorf1962adaptive,mahmud1989high,peng2009adaptive,feizi2010locally}. The essential difference is that in signal processing, measurements tend to be exact whereas in our setting a measurement just reveals the outcome of a single Bernoulli trial. 
Another popular related direction is \textit{online learning} or \textit{lifelong learning} during concept drift. Despite a large and growing body of work in this direction, including \citep{fontenla2013online,hoi2014libol,nallaperuma2019online,gomes2019machine,chen2019novel, nagabandi2018deep, hayes2020lifelong, chen2018lifelong, liu2017lifelong, hong2018lifelong}, this problem is by no means solved. Our setting assumes that after some time $T$, the model will eventually be retired for an updated one. It would be interesting to allow a policy to update $f$ based on the queried labels. Model \emph{robustness} is a related topic that focuses on designing $f$ such that accuracy does not fall during distribution drift \citep{zhao2019learning,lecue2020robust,li2018principled, goodfellow2018making, zhang2019building, shafique2020robust, miller2020effect}.

\vspace{-4pt}
\paragraph{Broader Impacts \& Limitations}  Anomaly detectors may be fragile. By robustifying the monitoring process, \textsc{MLDemon} can help organizations and stakeholders safely deploy ML models. \textsc{MLDemon}'s limitation is that it lacks the ability to go beyond alert generation. An interesting direction for future work is to incorporate downstream model repair and retraining together with monitoring.  

\vspace{-4pt}
\paragraph{Conclusion \& Future Directions} We pose  and analyze a novel formulation for studying automated ML deployment monitoring. Understanding the trade-off between expert attention and monitoring quality is of both research and practical interest. Our proposed policy comes with theoretical guarantees and performs favorably on empirical benchmarks. The potential impact of this work is that \textsc{MLDemon} could be used to improve the reliability and efficiency of ML systems in deployment.
Since this is a relatively new research direction, there are interesting directions for future work. We have assumed that experts respond to label requests instantly. In the future, we can allow the policy to incorporate labeling delay. Also, we have assumed that an expert label is as good as a ground-truth label. We can relax this assumption to allow for noisy expert labels. We could also let the policy more actively evaluate the apparent informativeness of the anomaly signal over time or even input an ensemble of different anomaly signals and learn which are most relevant at a given time. While \textsc{MLDemon} is robust even if the feature-based anomaly detector is not good, it is more powerful with an informative detector.
Improving the robustness of anomaly detectors for specific applications and domains is a promising area of research.

\subsection*{Acknowledgements}
We'd like to acknowledge V. Bagaria and B. He for useful conversations about this work. Also, we thank A. Grosskopf for helping with figure design and editorial help with the manuscript. Additionally, we thank the anonymous reviewers for helpful feedback. A.A.G. is supported by a Stanford Bio-X Fellowship. M.J.Z. is supported by NIH R01 MH115676. J.Z. is supported by NSF CAREER 1942926 and funding from the Chan-Zuckerberg Initiative and the Stanford AI Lab.

\bibliographystyle{apalike}  
\bibliography{references}


\clearpage
\appendix

\thispagestyle{empty}

\onecolumn \makesupplementtitle

\section{ IMPLEMENTATION DETAILS}

\subsection{Hyperparameter Selection for \textsc{MLDemon}}

First, we describe in more detail the hyperparameter selection for \textsc{MLDemon}. We delve into the quantitative laws that bind the hyperparameters; these ultimately enable \textsc{MLDemon} to automatically select some of the hyperparameters as discussed below.

In total, \textsc{MLDemon} has the following hyperparameters: $n, \alpha, \Delta, \epsilon, \rho$ and $q$. Of these, only two out of $n, \alpha, \Delta$ and $\epsilon$ are to be specified, with the exception of the pair $n,\epsilon$. Recall that $\rho$ specifies the threshold when in  a decision problem. Without loss of generality, we let $\rho \geq 1/2$ (if $\rho < 1/2$, simply reflect about $1/2$). For most of our analysis, the particular choice of $\rho$ only impacts the key quantities up to a constant factor, and thus we sometimes omit explicit reference to $\rho$ or carry out the analysis with $\rho \gets 1/2$.

Generally, it always makes sense to specify $n$ (the batch size of a the label queries) and $\alpha$ (specifies the ratio of the minimum waiting period in between label batches to the batch size). Based on these selections, \textsc{MLDemon} automatically chooses rate optimal pairs $(\epsilon, \Delta)$ such that if the drift is indeed $\Delta$-Lipschitz, \textsc{MLDemon} will guarantee expected monitoring risk lower than $\epsilon$. 

Another practical option is to specify $\epsilon$ and $\alpha$. In this case, \textsc{MLDemon} automatically chooses a batch size $n$ in order to guarantee $\epsilon$ and computes the best possible $\Delta$ for which it can guarantee $\epsilon$ given $n$ and $\alpha$.

In our theoretical analysis, $\Delta$, $q$ and $\rho$ are problem instances parameters. The threshold $\rho$ is known in practice since it is generally set by the system designer. Obviously, in practice, $\Delta$ might not be known (or may not even exist). However, sometimes, $\Delta$ can reasonably be upper bounded from historical trends, in which case one is also free to specify it. Although we do not formally address this here, the results in this paper should also generalize to the case in which the drift is $\Delta$-Lipschitz \emph{with high probability}.

As for $q$, it is of course not known in practice, but the choice $q \gets 1 - \epsilon$ has solid theoretical properties and is recommended.

\paragraph{Law for $n$ as a function of $\epsilon$:}
\begin{equation}
    n \gets \frac{9 \log (2\rho / \epsilon)}{2 \epsilon^2}
\end{equation}
\paragraph{Law for $\epsilon$ as a function of $n$:}
\begin{equation}
    \epsilon \gets 2\rho \cdot  \exp{\left( -\frac{1}{2}W\left(\frac{16n\rho^2}{9} \right)  \right)}
\end{equation}
where $W$ denotes the Lambert-W function \citep{veberivc2012lambert,valluri2000some}.
\paragraph{Law relating $\Delta$, $\epsilon$, and $\alpha$:}
\begin{equation}
\label{eqn:law_for_alpha_eps_delta}
    \Delta = \frac{\epsilon^3}{15 \alpha \log(2\rho/\epsilon)}
\end{equation}

Note that these laws are not arbitrary. They are constructed intentionally. Reading the proofs should illuminate them.

\subsection{Data Stream Details}
We describe each of the eight data streams in greater detail. All data sets are public and may be found in the references. All lengths for each data stream were determined by ensuring that the stream was long enough to capture interesting drift dynamics.

\subsubsection{Data Stream Construction}
\begin{enumerate}
     \item \texttt{SPAM-CORPUS}: We take the first $7400$ points from the data in the order that it comes in the data file. 
     
 
      \item \texttt{WEATHER-AUS}: For each random seed, we uniformly at random select of block of length $45,000$ from the data while preserving the \emph{chronological} the order of the data.

   
    \item \texttt{FACE-R}: We randomly subsample 400 individuals out of the data set that have at least 3 unmasked images and 1 masked image  to create a reference set. For these 400 individuals, we begin with a masking fraction of 10\%, and at some uniformly at random point in time (for each seed) we increase the masking fraction to 99\%. The total duration is 40,000.

    
\end{enumerate}

\subsubsection{Bootstrapping}

In order to get iterates for each data set, we generate the stream by bootstrap as follows. We block the data sequence into blocks of length $32$ and uniformly at random permute the data within each block. Because $ 32 \ll T$ this bootstrapping preserves the structure of the drift.

\subsection{Models}

\subsubsection{Logistic Regression}
For the logistic regression model, we used the default solver provided in the \texttt{scikit-learn} library \citep{pedregosa2011scikit}. As mentioned in the main text, we compute confidence scores in the usual way, meaning that we just take the logit value. Because these models are shallow, standard training routines produce fairly well calibrated models.

\subsubsection{Facial Recognition}
For the facial recognition system, we used the open-source model referenced in the main text. The model computes face embeddings given images. The embeddings are then used to compute a similarity score as described in the API. For any query image belonging to one of 400 individuals, the model looks for the best match among the 400 individuals by comparing the query image to each of the 3 reference images for each individual and taking an average similarity score. The highest average similarity score out of the 400 individuals is returned as the predicted matching individual. 

\section{MATHEMATICAL DETAILS AND PROOFS}
\normalsize
We begin with reviewing definitions and notations. We will then proceed with proofs for the results. We strongly recommend reading the proofs in the presented order.

\subsection{Definitions \& Notation}

\subsubsection{Problem Instances}
Although this we have defined the notion of a \emph{problem instance} throughout the main text, we briefly but formally revisit this here.

We consider sequences of distributions $\{P_t\}$ each over $(X,Y)$ that are $\Delta$-Lipschitz in the sense defined in the main text's problem formulation. Each problem instance has a fixed model $f$ and accuracy at time $t$ given by $\mu_t = \mathbf{Pr}[Y_t = f(X_t)]$. Additionally, each problem instance specifies a detection function $g$ that is parameterized by representation $E$ and statistical metric  $\mathsf{d}$ (as described in the main text). 

\begin{definition} \textbf{Space of all Distributions over a Measurable Space}

For a given measurable space $\Omega$, we let $\mathbb{P}(\Omega)$ denote the set of all probability distributions over $\Omega$. In the case that $\Omega$ is Euclidean, we assume the standard Borel $\sigma$-algebra.
\end{definition}

\begin{definition} \textbf{Anomaly Detector}

An anomaly detector (or anomaly detection signal) $g$ is pair $(E, \mathsf{d})$ such that $E:\mathcal{X} \rightarrow \mathbb{R}^s$ and $\mathsf{d}: \mathbb{P}(\mathbb{R}^s ) \times \mathbb{P}(\mathbb{R}^s) \rightarrow \mathbb{R}^{+}$. We let $\mathcal{G}$ denote the space of all such possible anomaly detectors.
\end{definition}

Additionally, we sometimes may assume the \emph{linear detection condition} (Def. \ref{def:linear_detection_condition}) which implicitly is a statement about some assumed distribution over problem statements. In other words, assuming the linear detection condition is like assuming that the problem instance itself is a random variable from a distribution that satisfies some additional constraints but is otherwise unknown.

\subsubsection{Initial Model Accuracy}

 We use the convention that $\mu_0$ is known from some held-out data. All policies can make use of this as their initial estimate.
\begin{definition} Accuracy at time $0$:
\begin{equation}
     \hat{\mu}_0 = \mu_0 
 \end{equation}
\end{definition}

\subsubsection{Monitoring Risk}
We first define the \emph{instantaneous monitoring risk}, $r$ which we distinguish here from the \emph{amortized monitoring risk} $R$ (defined in the main text). Instantaneous monitoring risk $r$ is the risk for a particular data point whereas $R$ is the amortized risk over the entire deployment.

\begin{definition} \textbf{Instantaneous Monitoring Risk}

We define the monitoring risks in MAE and hinge settings for a single data point in the stream below.

\begin{equation}
    r_\mathbf{mae}(\theta, \hat \theta) = |\theta - \hat \theta| 
\end{equation}

\begin{equation}
r_\mathbf{hinge}(\theta, \hat \theta; \rho) = \vert \rho - \theta\vert \left( \mathbf{1}\{\theta>\rho,\hat{\theta}<\rho \}  + \mathbf{1}\{\theta<\rho,\hat{\theta}>\rho \} \right)
\end{equation}

\end{definition}

As mentioned in the main text, we omit the subscript when the loss function is clear from context or is not relevant. In the context of our online problem, at time $t$ for policy $\pi$, we may generally infer that $\theta = \mu_t$,  $\hat\theta = \hat{\mu}_t$ and $\rho$ is fixed over time. In this case, we might use the shorthand $r^{\pi}(t)$, as below:

\begin{definition} \textbf{Amortized Monitoring Risk}

\begin{equation}
R^{\pi} = \frac{1}{T}\sum_{t=1}^T r^{\pi}(t)
\end{equation}

where $\pi$ is the policy and $R$ is the usual amortized monitoring risk term defined in Section 2.
\end{definition}

We may also omit the superscript $\pi$ when the policy is clear from context.

Also recall from Section 2 that we defined the query rate $Q$:

\begin{definition} \textbf{Amortized Query Rate}

\begin{equation}
Q^\pi = \frac{1}{T}\sum_{t=1}^T {a_t}
\end{equation}

where $a_t =1 $ if the policy queries a label at time $t$ and $a_t =0$ otherwise.
\end{definition}

\begin{definition} \textbf{Policy Loss}

\begin{equation}
\mathcal{L} = R + cQ
\end{equation}

for some $0 \leq c \leq 1$ that encodes trade-off between labeling cost and monitoring risk.
\end{definition}

\subsubsection{Policies}

We use the following abbreviations to formally denote the PQ, RR, and \textsc{MLDemon} policies: $\mathbf{PP}$, $\mathbf{RR}$, and $\mathbf{MLD}$, respectively. These three policies are defined in the main text, so we only briefly review them here.

\paragraph{PQ} We let $\mathbf{PQ}$ denote the PQ policy as defined in Section 3 of the main text. Recall that $\mathbf{PQ}$ can be parameterized by a particular query rate budget $B$ as defined in the main text. Alternatively, we can parameterize $\mathbf{PQ}$ by an upper bound of the worst-case risk tolerance $\epsilon$ such that $\mathbb{E}[R] \leq \epsilon$ in any problem instance. Using the theory we will presently develop, we can convert a risk tolerance $\epsilon$ into a constant average query rate given by $1/\alpha$ (based on $\alpha$ as computed in A.1). We use the same $\alpha$ for $\mathbf{PQ}$ as well as for \textsc{MLDemon}. The guaranteed risk tolerance $\epsilon$ implicitly depends on the Lipschitz constant $\Delta$, which we can assume to be known or upper bounded for the purposes of our mathematical analysis. For our asymptotic theory regarding PQ, we are thus implicitly using a hyperparameterization satisfying $B = \Theta(1/\alpha) = \widetilde{\Theta}(\Delta^{1/4})$ and $n= \widetilde{\Theta}(\Delta^{-1/2})$. By convention, if $\hat{\mu}_t$ is not updated at a given time $t$, then $\hat{\mu}_t \gets \hat{\mu}_{t-1}$. 

\paragraph{RR}  RR is defined in the main text. We write $\mathbf{RR}(\phi)$ to emphasize the dependence on a particular threshold hyperparameter $\phi \geq 0$. Recall that $\phi$ is set at time $0$ and fixed throughout deployment. When $\phi$ is omitted, the dependence is to be inferred. RR is the same for both decision and estimation problems.

\paragraph{MLD}

We let MLD denote the \textsc{MLDemon} policy. For our theoretical analysis, we shall parameterize  directly by $\epsilon$ and $\Delta$. In practice, we might use an estimate for $\Delta$ or just use $\alpha$ to parameterize MLD (as discussed in A.1). Notice that if $\Delta$ is not used as a parameter for MLD, our theoretical analysis has the alternative interpretation of defining an upper bound for $\Delta$ based on the hyperparameter conversions used by \textsc{MLDemon}.

We begin by reviewing and explicitly defining some of the key quantities used in \textsc{MLDemon}. We encourage the reader to revisit the code sketch (Alg. \ref{alg:mldemon}) in the main text if he or she should need a refresher on the algorithm. For the purposes of ensuring quantities internal to \textsc{MLDemon} are well-defined at all times, if they are not explicitly set or updated at a given time $t$, then assume the value from the previous time $t-1$ carries over. Quantities that are not explicitly initialized may be initialized arbitrarily.

Admittedly, some of the following definition may seem somewhat mysterious at first. As we develop our theory, it will become clear how each definition is used.

\begin{definition} (Decision Margin)
\label{def:dec_margin}

For decision problems, at time $t$, the \emph{decision margin} is given by $\ell_t$:

\begin{equation}
    \ell_t \gets \max\{ |\hat{\mu}_t - \rho| - \epsilon, 0\}
\end{equation} 

For estimation problems,
\begin{equation}
    \ell_t \gets 0
\end{equation} 

\end{definition}

\begin{definition} (Realized Bias Correction Term)
\label{def:realized_bias_correction}

Let $\tau$ denote the amount of time elapsed since \textsc{MLDemon}'s most recent query. At time $t$, the \emph{realized bias correction} is given by $\nu_t$:

\begin{equation}
    \nu_t \gets \Delta\cdot (\tau + (n+1)/2)
\end{equation} 

\end{definition}

The following definition makes use of quantities $G_t$ and $\partial_t^\mu$. These quantities are discussed in the main text and defined in Alg. \ref{alg:mldemon}.

\subsubsection{Confidence Intervals for \textsc{MLDemon}}
Of central importance to MLD is how we go about constructing the confidence intervals. We can think of $p_{\text{lbl}}$ as the likelihood that $\hat{\mu}_t$ is outside of the $\epsilon$-ball around $\mu_t$ according to the information in the label queries. We can think of $p_{\text{det}}$ as the likelihood that $\hat{\mu}_t$ is outside of the $\epsilon$-ball around $\mu_t$ according to the information in the anomaly detection signal. For decision problems, we can further increase the interval width to based on the decision margin. Using the linear detection condition, \textsc{MLDemon} can aggregate these two disparate sources of information with Bayesian inference.

While the construction of these confidence intervals are given Alg. \ref{alg:mldemon}, we review them here, beginning with $p_{\text{det}}$.

\begin{definition} (Data for Drift Estimation)
\label{def:data_for_drift_estimation}

At time $t$, the \emph{data for drift estimation} is set of comprised of pairs $(G_t, \partial_t^\mu)$. At the end of each batch of queries, $G_t$ and $\partial_t^\mu$ are added to the data set as new points. 
\end{definition}

\begin{definition} (Weight for Drift Estimation)
\label{def:weight_for_drift_estimation}

At time $t$, the \emph{weight for drift estimation}, denoted by the $\hat{w}_t$, is the OLS solution using the data for drift estimation to predict $\partial^\mu$ given $G$.
\end{definition}

Of course, we can apply constant time OLS updates rather than recomputing the OLS weight from scratch.

We now state Prop. \ref{prop:ols_confidence}. This is a standard result, so we simply refer the reader to an appropriate text.

\begin{proposition} (Prediction Interval for Ordinary Least Squares \citep{buteikis2019practical})
\label{prop:ols_confidence}

Assume $(X,Y) \sim \mathsf{P}$ follow a linear model with iid zero-mean Gaussian noise. Let $\mathsf{t}_m^{-1}$ denote the inverse CDF (i.e., quantile function) of the student-$\mathsf{t}$ distribution with $m$ degrees of freedom. Let $\mathsf{se}$ denote the standard error of the forecast for an OLS fit based on $N$ iid samples from $\mathsf{P}$. Let $\hat{Y}$ denote the OLS point estimate for $Y$ given $X$. Then, a $1-p$ \emph{prediction interval} is given by

$$\mathbf{Pr}[|Y - \hat{Y}| \geq \mathsf{se}\cdot \mathsf{t}_{N-2}^{-1}(1-p/2) ] \leq p$$

\end{proposition}

More details regarding Prop. \ref{prop:ols_confidence} can be found in \cite{buteikis2019practical}. Based on Prop. \ref{prop:ols_confidence}, we can easily derive a $1-p_{\text{det}}$ interval for the $\epsilon$-ball around $\mu_t$ in MLD.

\begin{proposition} (Detection-Based Confidence Interval)
\label{prop:label_conf_inter_det}

Let $\mathsf{t}_m$ denote the CDF of the student-$\mathsf{t}$ distribution with $m$ degrees of freedom. Let $\mathsf{se}_t$ denote the standard error of the forecast \citep{buteikis2019practical} for $\hat{w}$ at time $t$. Let $N$ be the number of data points in the dataset for drift estimation.

Then a $1-p_{\emph{det}}(t)$ confidence interval for $I(t)$ is given by:

\begin{equation}
       p_{\emph{det}}(t) =  2 -2\mathsf{t}_{N-2}\left(
      \frac{\epsilon + \ell_t -n\Delta}{\mathsf{se}_t} \right)
\end{equation} 

\begin{equation}
     I(t) = [\hat{\mu}_t - \epsilon  -\ell_t + n\Delta, \hat{\mu}_t + \epsilon +\ell_t - n\Delta] \cap [0,1]
\end{equation}

\end{proposition}

\begin{proof}
For the estimation case, this follows directly from Prop.\ref{prop:ols_confidence} by solving for $p$ in terms of the interval width. For the decision case, simply note that the policy is only penalized if the $\hat{\mu}_t$ is on the wrong side of $\rho$, meaning we can extend our confidence interval using the decision margin.
\end{proof}

Beyond this section, we generally omit the explicit time dependency for $p_{\text{lbl}}, p_{\text{det}}$ and interval $I$. We now turn our attention to discussing how we can derive $p_\text{lbl}$.

\begin{proposition}(Label-Based Confidence Interval)
\label{prop:label_conf_inter_lbl}

At time $t$, a $1-p_{\emph{lbl}}(t)$ confidence interval for $I(t)$ is given by:

\begin{equation}
       p_{\emph{lbl}}(t) = 1 - 2\exp\left(-2n(\epsilon - \nu_t + \ell_t - n\Delta)^2\right)
\end{equation}

\begin{equation}
      I(t) = [\hat{\mu}_t - \epsilon  -\ell_t + n\Delta, \hat{\mu}_t + \epsilon +\ell_t - n\Delta] \cap [0,1]
\end{equation} 
\end{proposition}

\begin{proof}
The estimation case follows directly from \ref{lemma:temp_hoeffding_for_delta_lip}. The decision case follows using the same reasoning as in the proof of \ref{prop:label_conf_inter_det}.
\end{proof}

\textsc{MLDemon} aggregates these two confidence intervals using standard Bayesian inference. The prior value of $q$ for the linear detection condition tells \textsc{MLDemon} how strongly to weigh the detection-based confidence interval against the label-based confidence interval.

\begin{proposition} (Bayesian Confidence Interval)
\label{prop:bayes_conf_inter}

Assume the linear detection condition with prior $q$. Then, at time $t$, a $1-p_t$ confidence interval for $I(t)$ is given by

\begin{equation}
       p_t = q\left(\frac{p_{\emph{lbl}}p_{\emph{det}}}{p_{\emph{lbl}}p_{\emph{det}} + (1-p_{\emph{lbl}})(1-p_{\emph{det}})} \right) + (1-q)p_{\emph{lbl}} 
\end{equation} 
\begin{equation}
      I(t) = [\hat{\mu}_t - \epsilon  -\ell_t + n\Delta, \hat{\mu}_t + \epsilon +\ell_t - n\Delta] \cap [0,1]
\end{equation} 
\end{proposition}
\begin{proof}
If the linear detection condition holds, we can directly combine the detection-based and label-based confidence intervals since they use the same interval $I$. If does not hold, we should only use the label-based confidence interval. Given the prior probability $q$ that the linear detection condition holds, we can use Bayes rule \citep{abu2012learning} to generate a confidence interval that probabilistically integrates between these two outcomes. 
\end{proof}

\subsection{Preliminary Results}

\subsubsection{Bounding Accuracy Drift in Absolute Value Based on Total Variation }

We begin with proving the claim from the introduction regarding the equivalence of a $\Delta$-Lipschitz bound in terms of accuracy drift and total variation between the distribution drift. Although Prop. \ref{prop:trivial_delta_lip} may not be immediately obvious for one unfamiliar with total variation distance on probability measures, the result is in fact trivial. To clarify notation, for probability measure $Q$ and event $\omega$ we let $Q(\omega) = \mathbb{E}_Q(\mathbf{1}\{\omega\})$.

\begin{proposition}
\label{prop:trivial_delta_lip}
Let $P$ and  $P'$ be two supervised learning tasks (formally,  distributions over $\mathcal{X} \times \mathcal{Y}$). Let $f : \mathcal{X} \rightarrow \mathcal{Y}$ be a model. If $d_{\text{TV}}(P,P') \leq \Delta$ then $|\mu - \mu'| \leq \Delta$ where $\mu = \mathbb{E}_P(\mathbf{1}\{f(X) = Y\})$ and $\mu' = \mathbb{E}_{P'}(\mathbf{1}\{f(X) = Y\})$.
\end{proposition}
\begin{proof}
Let $\mathbf{\Omega}$ be the sample space for distributions $P$ and $P'$. TV-distance has many equivalent definitions. One of them is  given in Eqn. \ref{eqn:tv_dist} below \citep{villani2008optimal}:

\begin{equation}
\label{eqn:tv_dist}
    d_{\text{TV}}(P,P') = \sup_{A \subset \Omega} |P(A) - P'(A)|
\end{equation}

\begin{equation}
   \geq |P(f(X) = Y) - P'(f(X) = Y)| = |\mu - \mu'|
\end{equation}

\end{proof}

\subsubsection{Adapting Hoeffding's Inequality for Lipschitz Sequences}

One of the key ingredients in many of the proofs is the following modification to Hoeffding's inequality  that enables us to use it to construct a confidence interval for the empirical mean of observed outcomes even though $P_t$ is drifting (rather than i.i.d. as is usual).

\begin{lemma}
\label{lemma:temp_hoeffding}
\textbf{Hoeffing's Inequality for Bernoulli Samples with  Bounded Bias}

Assume we have a random sample of $n$ Bernoulli trials such that each trials is biased by some small amount: $X_i \sim \mathbf{Bern}(p + \varepsilon_i)$.
Let $\bar X = \frac{1}{n}\sum_i X_i$ denote a sample mean. Let $\psi = \frac{1}{n} \sum_i |\varepsilon_i|$.

Then:

$$
\mathbf{Pr}( |\bar X - p| \geq \delta + \psi ) \leq 2\exp(-2n\delta^2)
$$

\end{lemma}

\begin{proof}
We can invoke the classical version of Hoeffding's \citep{hoeffding1994probability}:

\begin{equation}
\mathbf{Pr}(|\bar X - \mathbb{E}(\bar X)| \geq \delta) \leq 2 \exp(-2n\delta^2)
\end{equation}

Notice that $\mathbb{E}(\bar X) = p + \bar{\varepsilon}$. Plugging in below yields:

\begin{equation}
\mathbf{Pr}(|\bar X -  p + \bar{\varepsilon}| \geq \delta) \leq  2\exp(-2n\delta^2)
\end{equation}

Also, notice that due to the reverse triangle inequality \citep{sutherland2009introduction}:

\begin{equation}
|\bar X -  p + \bar{\varepsilon}| \geq ||\bar X -  p| - |\bar{\varepsilon}|| \geq |\bar X -  p| - |\bar{\varepsilon}| \geq|\bar X -  p| - \psi  
\end{equation}

Implying:

\begin{equation}
\mathbf{Pr}(|\bar X -  p | - \psi \geq \delta) \leq  \mathbf{Pr}( |\bar X -  p + \bar{\varepsilon}| \geq \delta)  \leq  2 \exp(-2n\delta^2) 
\end{equation}

Moving the $\psi$ to the other side of the inequality finishes the result:

\begin{equation}
\mathbf{Pr}(|\bar X -  p | \geq \delta + \psi)   \leq 2 \exp(-2n\delta^2) 
\end{equation}

\end{proof}


All of the work has already been in done above in Lem. \ref{lemma:temp_hoeffding},  but in order to make it more clear how it is applied to a Lipschitz sequence of distributions ${P_t}$ we also state Lem \ref{lemma:temp_hoeffding_for_delta_lip}.

\paragraph{Bias Correction Term} 
It will be convenient to explicitly define the bias correction term, $\psi$, from the lemma above. Notice that \textsc{MLDemon} explicitly keeps track of the bias correction term at time $t$, denoted by $\psi_t$ within the algorithm.

\begin{definition} (Bias Correction Term)

We denote the \textbf{bias correction term}, with $\psi$:

\begin{equation}
    \psi = \frac{1}{n}\sum_i |\epsilon_i|
\end{equation}

as defined in Lemma \ref{lemma:temp_hoeffding}.
\end{definition}

\begin{lemma}
\label{lemma:temp_hoeffding_for_delta_lip}
\textbf{Hoeffing's Inequality for Lipschitz Sequences}

Assume drift $\{P_t \} \in \mathbf{Lip}(\Delta)$ is $\Delta$-Lipschitz. Let $\mathcal{I}$ be any subset of the set of rounds for which policy $\pi$ has queried:

\begin{equation}
\mathcal{I} \subset \{i : C_i \geq 0 \}
\end{equation}

Let 

\begin{equation}
\overline C = \frac{1}{|\mathcal{I}|}\sum_{i\in \mathcal{I}} C_i
\end{equation}

denote a sample mean of observed outcomes for rounds in $\mathcal{I}$. 

Let 

\begin{equation}
    \psi = \frac{1}{|\mathcal{I}|}\sum_{i\in \mathcal{I}} |t-i|\Delta 
\end{equation}

be defined analogously to Lemma \ref{lemma:temp_hoeffding}.

Then:

\begin{equation}\label{eqn:hoeff_lip_seq}
\mathbf{Pr}( |\overline C - \mathbb{E}\{f(X_t) = Y_t \}| \geq \delta + \psi ) \leq 2\exp(-2n\delta^2)
\end{equation}

\end{lemma}

\begin{proof}
The result follows by setting $\varepsilon_i = |t-i|\Delta$ and applying Lemma \ref{lemma:temp_hoeffding}.
\end{proof}



\subsection{Zero-Noise Detection and Perfect Models}

It will prove fruitful to define and study a particular class of problem instance. 

\begin{definition} (Zero-Noise Linear Detection Instance)
We say a problem instance is a \emph{zero-noise linear} (ZL) instance if  the linear detection condition holds with $\mathcal{N} = 0$ for all $t$. 
\end{definition}

\begin{definition} (Zero-Noise Linear Detection Instance with Perfect Model)
We say a problem instance is a \emph{zero-noise linear with perfect model} (ZLPM) instance if it is ZL and additionally $\mu_t = 1$ for all $t$. 
\end{definition}

Consider a ZLPM instance. In this case, when using \textsc{MLDemon}, the standard error of the forecast for \textsc{MLDemon}'s linear model is 0: 

\begin{equation}
    \mathsf{se} = 0
\end{equation}

Based on this, we obtain perfect confidence surrounding our estimate: $p_\text{det} = 1$. 

Recall the update rule for setting \textsc{MLDemon}'s internal confidence intervals around $\hat{\mu}_t$:

\begin{equation}
p_t \gets q\left(\frac{p_{\text{lbl}}p_{\text{det}}}{p_{\text{lbl}}p_{\text{det}} + (1-p_{\text{lbl}})(1-p_{\text{det}})} \right) + (1-q)p_{\text{lbl}}
\end{equation}

As we shall see in Lemma \ref{lemma:finite_query_period}, the relationship between monitoring risk tolerance $\epsilon$ and linear detection prior $q$ determines \textsc{MLDemon}'s query period.

For now, we point out that ZLPM problem instances are deterministic and thus \textsc{MLDemon}'s behavior is deterministic in such instances.

\subsection{Proof of Lemma \ref{lemma:worst_case_tradeoff}}

 We study the average query rate required to achieve a worst-case monitoring risk, taken over all $\Delta$-Lipschitz drifts. Note that in the worst-case, the anomaly signal $\{G_t\}$ is uninformative and thus adaptivity with respect to the detection signal will not be helpful. The following results hold both for MAE loss and hinge loss. Lemma \ref{lemma:worst_case_tradeoff} has two parts. The first statement is about PQ whereas the second is for \textsc{MLDemon}.

We begin by proving Lem. \ref{lemma:worst_case_tradeoff} for PQ. Some constructions and equations derived in this proof will be useful in later proofs as well, which is why we start with this. We will return to Lem. \ref{lemma:worst_case_tradeoff} later on to prove it for \textsc{MLDemon} too.

\subsubsection{Lemma \ref{lemma:worst_case_tradeoff} for 
Periodic Querying}

\begin{lemma} 
\label{lemma:proof_for_PQ_monitor_risk} (Lemma \ref{lemma:worst_case_tradeoff} for PQ)

Let  $\mathbf{Lip}(\Delta)$ be the class of $\Delta$-Lipschitz drifts. Assume $\Delta \leq \frac{\epsilon^3}{10\log(2/\epsilon)}$. For both estimation and decision problems (using MAE and hinge loss), PQ achieves a worst-case expected monitoring risk of $\epsilon$ with a query rate of $O\left(\frac{\Delta \log(1/\epsilon)}{\epsilon^3}\right)$:

\end{lemma}
\begin{proof}

Because $r_\mathbf{mae}(t) \geq r_{\mathbf{hinge}}(t)$ for all $t$, it will be sufficient to prove the result for the estimation case (doing so directly implies the decision case).

By construction, the amortized query complexity for PQ (Alg. \ref{alg:PQ}) is $\Theta(\frac{1}{\alpha})$. This query rate indeed satisfies the query rate condition. This holds for any choice of $\{P_t\}$ because PQ is an open-loop policy.

\begin{equation}
\label{eqn:alpha_2_eps}
Q^{\mathbf{PQ}} \leq \Theta\left(\frac{1}{\alpha}\right) = O\left(\frac{\Delta \log(1/\epsilon)}{\epsilon^3}\right)
\end{equation}

It remains to verify that the choice of $n$ and $\alpha$ results in a worst-case expected warning risk of $\epsilon$.

Consider Lemma \ref{lemma:temp_hoeffding_for_delta_lip}. Based on Lemma \ref{lemma:temp_hoeffding_for_delta_lip}, imagine we are applying Eqn. \ref{eqn:hoeff_lip_seq} at each point in time $t$ with quantity $\overline{C}$ being used as our point estimate $\hat{\mu}$. If, for all time: 

\begin{equation}
\label{eqn:main_hoeff_ineq}
    \psi + \delta \leq \epsilon \textbf{ and } 2\exp({-2n\delta^2}) \leq \epsilon
\end{equation}

Then it follows from  Eqn. \ref{eqn:hoeff_lip_seq} that PQ attains monitoring risk $\epsilon$ because we the event $|\hat{\mu}_t - \mu_t| \geq \epsilon$ occurs with probability less than $\epsilon$ (and, of course, $|\hat{\mu}_t - \mu_t| \leq 1$ always).

Fixing $\delta \gets \epsilon/3$, it is easy to verify that $2\exp({-2n\delta^2}) \leq \epsilon$. 

Recall that $n$ is fixed:

\begin{equation}
n = \frac{9\log(2/\epsilon)}{2\epsilon^2}
\end{equation}

Plugging in the above $n,\delta$ produces:

\begin{equation}
2\exp({-2n\delta^2}) = \epsilon
\end{equation}

It remains to verify the first inequality, $\psi + \delta \leq \epsilon$, at all $t$. This is slightly more involved, as $\psi$ is not constant over time. We ask ourselves, what is the worst-case $\psi$ that would be possible in Eqn. \ref{eqn:hoeff_lip_seq} when using PQ? Well, the PQ policy specifies a query batch of size $n$, and maintains the empirical accuracy from this batch as the point estimate for the next $(\alpha + 1)n$ rounds in the stream. Thus, $\psi$ is largest when precisely when the policy is one query away from completing a batch. At this point in time, because the policy has not yet updated $\hat{\mu}$ because it only does so at the end of the batch once all $n$ queries have been made. Thus, we are using an estimate that is the empirical mean of $n$ label queries such that this batch was started $n(\alpha + 1) -1 $ iterations ago and was completed $n\alpha - 1 $ iterations ago. 

The maximal value possible for $\psi$ is hence:

\begin{equation}
    \max_{t} \{ \psi \} = \frac{\Delta}{n}\sum_{i=n\alpha - 1}^{n(\alpha +1) -1} i
\end{equation}
It is easy to upper bound this sum as follows:
\begin{equation}
\label{eqn:upper_bound_psi}
 < \frac{\Delta}{n}\sum_{i=n\alpha}^{n(\alpha +1)} i = \frac{\Delta}{n}(n^2\alpha + \sum_{i=1}^{n} i) = \Delta n\alpha + \Delta(n+1)/2
\end{equation}

Recall $\alpha$:

\begin{equation}
     \alpha = \frac{\epsilon^3}{15\Delta\log(2/\epsilon)} 
\end{equation}

Plugging in $\alpha, n$ into $\max\{\psi\}$, along with the upper bound for $\Delta \leq \frac{\epsilon^3}{10\log(2/\epsilon)}$ yields:

\begin{equation}
\label{eqn:ineq_for_delta_n_alpha_to_eps}
  \Delta n \alpha  < \epsilon/3
\end{equation}

\begin{equation}
    \Delta(n+1)/2 \leq \frac{9}{40} \epsilon  + \frac{1}{20}\epsilon^3 < \epsilon/3
\end{equation}

Together, these inequalities imply:

\begin{equation}
\label{eqn:psi_up_bound}
     \psi < \Delta n \alpha + \Delta(n+1)/2  < 2\epsilon/3
\end{equation}

Recalling that we fixed $\delta$:

\begin{equation}
  \delta \gets \epsilon/3
\end{equation}

We conclude,

\begin{equation}
  \psi + \delta \leq \epsilon \text{ } \forall t
\end{equation}

Thus, by application Lemma \ref{lemma:temp_hoeffding_for_delta_lip}, for all $t$:

\begin{equation}
\label{eqn:end_of_C20}
 \mathbb{E}[r^{\mathbf{PQ}}(t)] \leq \epsilon \text{ } \forall t
\end{equation}

from which the amortized result follows via the linearity of expectation.

\end{proof}

\subsubsection{Lemma \ref{lemma:worst_case_tradeoff} for \textsc{MLDemon}}
To begin, we clarify an important distinction regarding MLD's monitoring risk in the following remark. 

\begin{remark}
\label{remark:mld_risk_2_eps}
Under the conditions of Thm. \ref{thm:worst_case_regret_main}, \textsc{MLDemon} achieves a worst-case expected monitoring risk of $2\epsilon$. Additionally, if $(E,\mathsf{d})$ satisfies the linear detection condition with probability $q$, then \textsc{MLDemon} achieves worst-case expected monitoring risk of $\epsilon$.
\end{remark}

Notice that based on Remark \ref{remark:mld_risk_2_eps}, the minimax rate for $\mathcal{L}$ remains the same regardless of if we admit the linear detection condition. Another interpretation is that when using \textsc{MLDemon} without wanting to admit the linear detection condition, one can halve the effective monitoring risk hyperparameter in order to get an exact guarantee.

\begin{lemma} 
\label{lemma:finite_query_period} 

Under the conditions of Thm. \ref{thm:worst_case_regret_main}, if $q < 1 - \epsilon$, then \textsc{MLDemon}'s query period is always finite. 

\end{lemma}
\begin{proof}

  The key to understanding \textsc{MLDemon} is to understand how the confidence intervals are computed. At each time $t$, \textsc{MLDemon} produces a confidence $p_t$ that $|\mu_t - \hat{\mu}_t| \leq \epsilon - n\Delta$. If $p_t \geq 1 - \epsilon$, then \textsc{MLDemon} can guarantee that a monitoring risk of $\epsilon$ by virtue of the confidence interval. The $n\Delta$ correction is needed to account for the $n$ points that go by while the next batch of labels is being collected.
  
  \textsc{MLDemon} assumes the linear detection condition at a prior probability $q$. As a result, \textsc{MLDemon} makes use of Bayes rule to combine the independent confidence intervals $p_{\text{lbl}}$ (from the label batches) and $p_{\text{det}}$ (from the anomaly detector). The dependence on $t$ for these two quantities is omitted.  
  
  From Bayes rule it follows:
  
  \begin{equation}
  \label{eqn:p_t_bayes}
      p_t = q\left(\frac{p_{\text{lbl}}p_{\text{det}}}{p_{\text{lbl}}p_{\text{det}} + (1-p_{\text{lbl}})(1-p_{\text{det}})} \right) + (1-q)p_{\text{lbl}} 
  \end{equation}
  
  Where the first term, multiplied by factor $q$, comes from the joint interval produced by two independent intervals, and the second term, multiplied by factor $1-q$ is the result of the possibility that the linear detection condition does not hold, in which case we only make use of the information from the labels, $p_{\text{lbl}}$. 
  
Label information will eventually become stale in time if no new batches are acquired. More concretely, assume there exists $\tau $ such that $a_t = 0$ for all $ t>\tau$, then 

\begin{equation}
    \lim_{t\rightarrow \infty} p_{\text{lbl}} = 0
\end{equation}
  
 Based on Eqn. \ref{eqn:p_t_bayes}, we can see that when $q < 1- \epsilon$
 \begin{equation}
     \lim_{t\rightarrow \infty} p_{t} < 1 - \epsilon
\end{equation}

Because $p_t$ is asymptotically upper bounded by $1 - \epsilon$, we can conclude that no such $\tau$ exists such that $a_t = 0$ for all $ t>\tau$. Otherwise, it would contradict  \textsc{MLDemon}'s contract that only delays a label batch if $p_t \geq 1 - \epsilon$.

Thus, the condition $q < 1- \epsilon$ is sufficient to establish a finite query period for \textsc{MLDemon}.
\end{proof}

After establishing that the query period is finite, the next step is to upper and lower bound the query period.

\begin{definition} (Maximal period extension)

We define \textsc{MLDemon}'s \emph{maximal period extension} as the largest possible increase in query period over all problem instances. \footnote{Recall that the query period for \textsc{MLDemon} is never shorter than that of PQ.}
\end{definition}

Note that the maximal period extension can vary significantly based on if we are in an estimation problem or decision problem. For the proof of Lemma \ref{lemma:worst_case_tradeoff}, we will primarily focus on the maximal period extension for estimation problems. The maximal period extension for decision problems turns out to be less relevant for Lemma \ref{lemma:worst_case_tradeoff}, but it does come up in the proof of Lemma \ref{lemma:avg_q_decision}.

\begin{lemma} 
\label{lemma:bounds_for_query_period_mld} 

 Assume $\Delta \leq \frac{\epsilon^3}{10\log(2/\epsilon)}$ and $q < 1 - \epsilon$. Under the conditions of Thm. \ref{thm:worst_case_regret_main}, for estimation problems (using MAE loss), \textsc{MLDemon}'s maximal period extension, denoted $\mathfrak{n}_{\max}$, is bounded by:

 $$ 1/3 \leq \frac{\Delta \mathfrak{n}_{\max}}{\epsilon} \leq 1$$

\end{lemma}
\begin{proof}

In this proof, our task is to bound $\mathfrak{n}_{\max}$, which denotes the largest possible $\mathfrak{n} \in \{ 0, \dots, \mathfrak{n}_{\max}\}$ that \textsc{MLDemon} would ever allow in any problem instance. 

  It will be helpful to recall notion and definition of bias correction $\psi$ from Lemmas \ref{lemma:temp_hoeffding_for_delta_lip} and \ref{lemma:proof_for_PQ_monitor_risk}. 

We aim to quantify the longest possible query period. At any given time $t$, recall from Alg. \ref{alg:mldemon} that $\tau(t)$ is the time since the end of the most recent batch of label queries. If $\tau \geq n \alpha$, then the current period extension is given by

\begin{equation}
    \mathfrak{n}(t) = \tau(t) - n \alpha
\end{equation}

Henceforth we omit the explicit time dependence. Obviously, $\mathfrak{n}_{\max} > 0$, so we will assume that $\mathfrak{n}(t) \geq 1$ at this particular time $t$.

 To account for the increase in bias correction required as the time elapsed since the most recent query batch grows, we can modify the upper bound to $\psi$ in Ineq. (\ref{eqn:upper_bound_psi}) as follows:



\begin{equation}
   \psi \leq  \Delta n \alpha  +  \mathfrak{n}\Delta + \Delta(n+1)/2 = \Delta(\tau + (n+1)/2) 
\end{equation}

Notice that this upper bound is precisely the realized bias correction $\nu_t$:

\begin{equation}
\nu_t =  \Delta(\tau + (n+1)/2) 
\end{equation}

Although, \textsc{MLDemon} does not explicitly set a value for $\delta$, we can actually set a virtual $\delta$ as follows:

\begin{equation}
   \delta = \epsilon -  \nu_t 
\end{equation}

In which case we can see that the conditions in \ref{eqn:main_hoeff_ineq} would become:

\begin{equation}
\label{eqn:mld_conditions_main}
    \nu_t + \delta \leq \epsilon \textbf{ and } 2\exp({-2n\delta^2}) \leq \epsilon
\end{equation}

The query conditions in Alg. \ref{alg:mldemon} immediately follow with $\delta \gets \epsilon -  \nu_t -n\Delta$ when we evaluate that $\ell_t \gets 0$ in the estimation case and we also provide the $n\Delta$ buffer term in order to ensure that we will be able to complete a query batch without violating the inequality.

Recall Ineq. ($\ref{eqn:psi_up_bound}$): $\psi \leq 2\epsilon/3$. With that in mind, we assert the following:

\begin{equation}
    \nu_t \leq \epsilon
\end{equation}

Due to the non-negativity of $\delta$ and that $\nu_t + \delta \leq \epsilon$. Combined with (1) the non-negativity of $\psi$ and (2) the upper bound $\psi \leq 2\epsilon/3$, we can determine that:

\begin{equation}
    \epsilon/3 \leq  \nu_t - \psi \leq \epsilon
\end{equation}

Dividing by $\epsilon$ and substituting $\nu_t - \psi = \Delta \mathfrak{n}$ yields the result.
\end{proof}

\begin{lemma} 
\label{lemma:proof_for_MLD_monitor_risk} (Lemma \ref{lemma:worst_case_tradeoff} for $\mathbf{MLD}$ under MAE Loss)

Let  $\mathbf{Lip}(\Delta)$ be the class of $\Delta$-Lipschitz drifts. Assume $\Delta \leq \frac{\epsilon^3}{10\log(2/\epsilon)}$ and $q < 1 - \epsilon$. Under the conditions of Thm. \ref{thm:worst_case_regret_main}, for estimation problems (using MAE loss), MLD achieves a worst-case expected monitoring risk of $2\epsilon$ with a query rate of $\widetilde{O}\left(\frac{\Delta}{\epsilon^3}\right)$:

\end{lemma}

\begin{lemma} 
\label{lemma:proof_for_MLD_monitor_risk} (Lemma \ref{lemma:worst_case_tradeoff} for $\mathbf{MLD}$ under MAE Loss)

Let  $\mathbf{Lip}(\Delta)$ be the class of $\Delta$-Lipschitz drifts. Assume $\Delta \leq \frac{\epsilon^3}{10\log(2/\epsilon)}$ and $q < 1 - \epsilon$. Under the conditions of Thm. \ref{thm:worst_case_regret_main}, for estimation problems (using MAE loss), MLD achieves a worst-case expected monitoring risk of $2\epsilon$ with a query rate of $\widetilde{O}\left(\frac{\Delta}{\epsilon^3}\right)$:

\end{lemma}
\begin{proof}

First, we analyze the query rate. Strictly speaking, it is trivial to see that MLD never increases the query rate compared to PQ. So, it is sufficient to bound the possible degradation in monitoring risk to prove this lemma. However, in order to built a bit of intuition, we will take a brief detour to show how our bounds on the maximal query period extension convert to bounds on the asymptotic query rate. Afterwards, we analyze the monitoring risk.

With upper and lower bounds on the maximal query period extension (Lemma \ref{lemma:bounds_for_query_period_mld}), we can establish that: \begin{equation}
\label{eqn:start_of_Q_analysis}
    \mathfrak{n}_{\max} = \Theta(\epsilon/\Delta)
\end{equation}

Recall that in ZPLM problem instances, $p_{\text{det}}=1$ for all $t$. This implies that \textsc{MLDemon}'s query period is always as long as possible. Let $P_{\text{ZPLM}}$ denote a ZPLM instance and we know:

\begin{equation}
    P_{\text{ZPLM}} \in \argmin_{P \in \mathbf{Lip}(\Delta)} \{ \mathbb{E}_P[ Q^{\mathbf{MLD}}]\}
\end{equation}

Furthermore, in a ZPLM problem instance outcomes and behavior are deterministic, so we know:

\begin{equation}
\mathbb{E}_{P_{\text{ZPLM}}}[ Q^{\mathbf{MLD}}] =  \frac{n}{(\alpha + 1)n + \mathfrak{n}_{\max}} 
\end{equation}



And therefore, the min query rate\footnote{Keep in mind that query rate and query period are reciprocal, so the minimal query rate is the maximal query period.} is given by:

\begin{equation}
\label{eqn:Q_mld_formula_full}
    \inf_{P \in \mathbf{Lip}(\Delta)} \{ \mathbb{E} Q^{\mathbf{MLD}} \} = \frac{n}{(\alpha + 1)n + \mathfrak{n}_{\max}} 
\end{equation}
\begin{equation}
\label{eqn:Q_mld_formula}
    = \frac{n}{(\alpha + 1)n + \Theta(\epsilon/\Delta)}
\end{equation}

Notice the following asymptotic conversions between variables:
\begin{equation}
    \alpha = \widetilde{\Theta}(\Delta^{-1/4})
\end{equation}
\begin{equation}
    \epsilon = \widetilde{\Theta}(\Delta^{1/4})
\end{equation}
\begin{equation}
    n = \widetilde{\Theta}(\Delta^{-1/2})
\end{equation}

Using the above conversions in Eqn. \ref{eqn:Q_mld_formula} yields:

\begin{equation}
    \inf_{P \in \mathbf{Lip}(\Delta)} \{ \mathbb{E} Q^{\mathbf{MLD}} \} = \frac{1}{1 + \widetilde{\Theta}(\Delta^{-1/4})} 
\end{equation}
\begin{equation}
\label{eqn:end_of_Q_analysis}
    = \widetilde{\Theta}(\Delta^{1/4}) = \widetilde{\Theta}(\Delta/\epsilon^3)
\end{equation}

The key insight from the above analysis is that $\mathfrak{n}_{\max} = \Theta(\alpha n) \sim \Delta^{-3/4}$, from which we can observe that the period extensions in $\mathbf{MLD}$ do not affect the minimax rates from an asymptotic perspective. 

To complete the proof, we must now turn our attention to the the worst-case expected monitoring risk, $\mathbb{E}\left[R^{\mathbf{MLD}}\right]$. However, given that the rates are invariant with respect to the period extensions, we should expect that the worst-case expected monitoring risk increases by at most a constant factor due to the period extension.

Recall the conditions from (\ref{eqn:mld_conditions_main}):

\begin{equation}
    \nu_t + \delta \leq \epsilon \textbf{ and } 2\exp({-2n\delta^2}) \leq \epsilon
\end{equation}

The second condition does not depend on the query period.\footnote{Indeed, the second condition in (\ref{eqn:mld_conditions_main}) is met when the initial estimate upon immediately completing a batch of label queries is sharp enough.} Rather, it is the first condition that could break down if $\nu_t$ grows too much as a result of too long a query period. However, we have Lemma \ref{lemma:bounds_for_query_period_mld} at hand to upper bound $\nu_t$ through $\mathfrak{n}_{\max}$.

By definition,
\begin{equation}
  \nu_t  \leq \psi + \Delta\mathfrak{n}_{\max}.
\end{equation}
    
And with Lemma \ref{lemma:bounds_for_query_period_mld}:
\begin{equation}
   \Delta\mathfrak{n}_{\max} \leq \epsilon
\end{equation}
    
Therefore,

\begin{equation}
\nu_t + \delta \leq \psi + \Delta\mathfrak{n}_{\max} + \delta \leq \psi + \epsilon + \delta \leq 2\epsilon
\end{equation}
    
From this, it follows that 
\begin{equation}
      \sup_{P \in \mathbf{Lip}(\Delta)} \{ \mathbb{E}_P [R^{\mathbf{MLD}}] \} \leq 2\epsilon = O(\epsilon)
\end{equation}

\end{proof}

\begin{lemma} 
\label{lemma:proof_for_MLD_monitor_risk_hinge} (Lemma \ref{lemma:worst_case_tradeoff} for $\mathbf{MLD}$ under Hinge Loss)

Let  $\mathbf{Lip}(\Delta)$ be the class of $\Delta$-Lipschitz drifts. Assume $\Delta \leq \frac{\epsilon^3}{10\log(2/\epsilon)}$ and $q < 1 - \epsilon$. Under the conditions of Thm. \ref{thm:worst_case_regret_main}, for decision problems (using hinge loss), MLD achieves a worst-case expected monitoring risk of $2\epsilon$ with a query rate of $\widetilde{O}\left(\frac{\Delta}{\epsilon^3}\right)$:

\end{lemma}
\begin{proof}

For this proof we must show that MLD does achieve a worst-case expected monitoring risk of $O(\epsilon)$ for decision problems. Of course, the max query rate for decision problems is the same as for estimation problems.

The possible issue with MLD for decision problems is that MLD is too aggressive with extending the query period based on the decision margin, resulting in a monitoring risk that is unacceptable.

However, it is not difficult to see that this is not the case. It is straightforward to note that the decision margin is the appropriate interval width for decision problems. Next, consider the following variant on  the conditions in \ref{eqn:main_hoeff_ineq}:

\begin{equation}
    \nu_t + \delta \leq \ell_t + 2\epsilon \textbf{ and } 2\exp({-2n\delta^2}) \leq \epsilon
\end{equation}

The factor of $2$ for the $\epsilon$ comes from Lemma \ref{lemma:proof_for_MLD_monitor_risk}. These are the conditions which must always hold true for MLD to maintain the monitoring risk guarantee.

Setting $\delta  =  2\epsilon - \nu_t + \ell_t - n\Delta$ is sufficient to satisfy the conditions. By construction MLD, only extends the query period as long as it can satisfy these conditions (with the $n\Delta$ correction as a buffer to give time to complete the next query batch).

\end{proof}

We can combine Lemma \ref{lemma:proof_for_MLD_monitor_risk} and Lemma \ref{lemma:proof_for_MLD_monitor_risk_hinge} to prove Lemma \ref{lemma:worst_case_tradeoff} from the main text. We now turn our attention to Thm. \ref{thm:worst_case_regret_main}, which shall make use of Lemma \ref{lemma:worst_case_tradeoff} in parts (i) and (ii).

\subsection{Proof of Theorem \ref{thm:worst_case_regret_main}}
\begin{theorem} (Theorem 4.1)
Let $\mathcal{P} = \mathbf{Lip}(\Delta)$ be the set of $\Delta$-Lipschitz drifts and let $\Pi$ be the space of deployment monitoring policies. On both estimation problems with MAE risk and decision problems with hinge risk, for any model $f$ and anomaly detector $g$, the following (i) - (iv) hold.

(i) PQ has a worst-case expected loss $$\sup_{P \in \mathcal{P}} \mathbb{E}_{P}[\mathcal{L}^{\mathbf{PQ}}_{g}] = \widetilde{O}\big( \Delta^{1/4}\big)$$ 

(ii) When $0 \leq q < 1$ is constant and $\epsilon = \Theta(\Delta^{1/4})$,  \textsc{MLDemon} has a worst-case expected loss $$\sup_{P \in \mathcal{P}} \mathbb{E}_{P}[\mathcal{L}^{\mathbf{MLD}}_{g}] = \widetilde{O}\big( \Delta^{1/4}\big)$$

\normalsize
(iii) RR has a worst-case expected loss $$\sup_{P \in \mathcal{P}} \mathbb{E}_{P}[\mathcal{L}^{\mathbf{RR}}_{g}] = \Theta(1)$$

(iv) No policy can achieve a better worse-case expected loss than \textsc{MLDemon} and PQ: $$\inf_{\pi \in \Pi} \text{ } \sup_{P \in \mathcal{P}} \mathbb{E}_{P}[\mathcal{L}^{\pi}_{g}] = \Omega\left( \Delta^{1/4}\right)$$

\end{theorem}

\subsubsection{Part (i)}

\begin{lemma} (Theorem 4.1.i)
\label{lemma:PQ_final_step}
   PQ has a worst-case expected loss $\sup_{P \in \mathcal{P}} \mathbb{E}_{P}[\mathcal{L}^{\mathbf{PQ}}_{g}] = \widetilde{O}\big( \Delta^{1/4}\big)$
\end{lemma}

\begin{proof}

 \begin{equation}
 \sup_{(g,P) \in \mathcal{G} \times \mathbf{Lip}(\mathbf{\Delta})} \mathbb{E}_{P}[\mathcal{L}^{\pi}_{g}] = \sup_{(g,P) \in \mathcal{G} \times \mathbf{Lip}(\mathbf{\Delta})} \mathbb{E}_{P}[R^{\pi}_{g} + cQ^{\pi}_{g} ] 
 \end{equation}
 
 \begin{equation}
     \leq \sup_{(g,P) \in \mathcal{G} \times \mathbf{Lip}(\mathbf{\Delta})} \mathbb{E}_{P}[R^{\pi}_{g}] + \sup_{(g,P) \in \mathcal{G} \times \mathbf{Lip}(\mathbf{\Delta})} \mathbb{E}_{P}[Q^{\pi}_{g}] 
 \end{equation}
 
 Because $\pi = \mathbf{PQ}(\epsilon)$ we can apply Lemma \ref{lemma:worst_case_tradeoff}:
 
  \begin{equation}
 = \epsilon + \widetilde{O}(\Delta / \epsilon^3)
 \end{equation}
 
 Risk tolerance $\epsilon$ is a user-specified parameter. Setting $\epsilon = \Theta(\Delta^{1/4})$ yields:
 
   \begin{equation}
 \epsilon + \widetilde{O}(\Delta / \epsilon^3) =  \widetilde{O}(\epsilon) =  \widetilde{O}(\Delta^{1/4})
 \end{equation}
 
 which completes the proof. 
\end{proof}

\subsubsection{Part (ii)}

\begin{lemma} (Theorem 4.1.ii)
   When $0 \leq q < 1$ is constant and $\epsilon = \Theta(\Delta^{1/4})$,  \textsc{MLDemon} has a worst-case expected loss $\sup_{P \in \mathcal{P}} \mathbb{E}_{P}[\mathcal{L}^{\mathbf{MLD}}_{g}] = \widetilde{O}\big( \Delta^{1/4}\big)$
\end{lemma}
\begin{proof}
See the preceding proof for Part (i). We follow the same argument, except that 
 $\pi = \mathbf{MLD}$. Given that Lemma \ref{lemma:worst_case_tradeoff} applies to both \textbf{MLD} and \textbf{PQ} under the assumptions, the proof from Part (i) also applies to Part (ii).
\end{proof}

\subsubsection{Part (iii)}

 We can contrast the rates from Parts (i) and (ii) with the minimax rate for RR. Lemma \ref{lemma:part_ii_of_minimax_theorem_main}  follows from the fact that in the worst-case the anomaly signal is poorly calibrated. Either the model accuracy drifts without alerting the detector or the policy will spuriously query too often.

\begin{lemma} (Theorem 4.1.iii)
 \label{lemma:part_ii_of_minimax_theorem_main}
For any initial distribution $(X,Y) \sim P_0$ The worst-case expected regret of RR is
 
 $$  \inf_{\phi} \text{ }  \sup_{P} \Big(  \mathbb{E}_P \big[\mathcal{L}^{\mathbf{R}(\phi)} \big] \Big) \geq  \mathbf{min} \{1-\rho, c\}$$

 \end{lemma}
 
 \begin{proof}
 
  At a high-level, the proof idea is that there always exists $P_t$ that can make the $\mathbf{R}(\phi)$ policy either query too much or too little, regardless of what the original $P_0$ is.
 
 For any head $\{x_1,...,x_{\tau-1}\}$ of any length $\tau$ and any $\chi \in \mathcal{X}$ the \textbf{constant} tail $\{\chi,\chi,\chi,...\} \in \{ \chi \}^{\infty}$ results in a constant anomaly signal $G_t = C$. This follows from the fact that feature data going into the detection windows is constant. If $C \geq \phi$, then RR is constantly querying, meaning that both the detection windows are constantly repopulated with the same data $(\chi, \chi, ..., 
 \chi)$ and if  $C < \phi$, then RR will never query again because one detection window will be held fixed and the second is constantly repopulated with the same data, producing the same $G_t$ for all time.

 However, recall that the result should hold for all possible initial distributions $P_0$. This will not prove to be a major obstacle though. 
 
 Below, we define distribution $P'$ (parameterized by $\chi \in \mathcal{X}$) over $\mathcal{X} \times \mathcal{Y}$ in terms of the marginal over $X$ and the conditional for $Y|X$.
 
 \begin{equation}
     P_{X}'(X = \chi) = 1,\text{ } P'_{Y|X} = (P_0)_{Y|X}    
 \end{equation}
 
 Thus, $P'$ is a point mass at $\chi$ while holding the same conditional as $P_0$. Note that $d_{\text{TV}}(P_0, P') \leq 1$ (this holds for any two distributions by definition of TV-distance). There exists a $\Delta$-Lipschitz sequence head $\{P_0,..., P_{\mathbf{ceil}(1/\Delta)} \}$ of length $\mathbf{ceil}(1/\Delta)$ such that $P_{\mathbf{ceil}(1/\Delta)} = P'$. For example, the head given by the following sequence of mixtures:
 
 \begin{equation}
     P_j = (j/\mathbf{ceil}(1/\Delta)) P_0 + (1-j/\mathbf{ceil}(1/\Delta))P'
 \end{equation}

For reasons to be made apparent later, we pad this sequence head with a buffer of length $m$ of repeating $P'$. Thus our head becomes $\{P_0,..., P_{\mathbf{ceil}(1/\Delta)},...,P_{m+\mathbf{ceil}(1/\Delta)} \}$ where $P_t = P'$ if $ \mathbf{ceil}(1/\Delta) \leq t \leq m+\mathbf{ceil}(1/\Delta)$.

We now turn our attention to constructing the tail of the sequence. We begin by defining $P''$ and letting $P_{m+\mathbf{ceil}(2/\Delta)} = P''$. For $P''$,  hold $X_t$ concentrated as a point mass on $\chi$. Thus it is sufficient to define $P''_{Y|X=\chi}(y)$. Let $V \sim \mathbf{Bern}(1/2)$.

 \[ P''_{Y|X=\chi}(y) = \begin{cases} 
    f(\chi) & \textbf{if } V = 1    \\
    y' \in  \{ \tilde{y} : \tilde{y} \neq f(\chi), \tilde{y} \in  \mathcal{Y} \}  & \textbf{if } V = 0\\
   \end{cases}
\]

As before, there must exists some sequence head $\{P_0,...,P_{\mathbf{ceil}(1/\Delta)},...,P_{m+\mathbf{ceil}(1/\Delta)},...,P_{m+\mathbf{ceil}(2/\Delta)} \}$ such that $P_{\mathbf{ceil}(1/\Delta)} = P'$ and $P_{\mathbf{ceil}(2/\Delta)} = P''$. Beyond time $t = \mathbf{ceil}(2/\Delta)$ we keep the sequence constant at distribution $P''$ such that the final sequence is given by 

\begin{equation}
    \{ P_0,...,P',...,P',...,P'',P'',P'',... \} 
\end{equation}    

  $$      \text{ where } P_t = P' \text { for }  \mathbf{ceil}(1/\Delta)\leq  t \leq m+\mathbf{ceil}(1/\Delta) $$
  and 
$$ P_t = P'' \text { for } t \geq {m+\mathbf{ceil}(2/\Delta)} $$

 And the intermediate length $\mathbf{ceil}(1/\Delta)$ segments
 
 $$
  \{P_1,...,P_{\mathbf{ceil}(1/\Delta)-1}\}$$
  
  $$\{P_{m+\mathbf{ceil}(1/\Delta)+1},...,P_{m+\mathbf{ceil}(2/\Delta)-1}\}
$$
are guaranteed to exist within the $\Delta$-Lipschitz constraint.

We can conclude that there exists a $\Delta$-Lipschitz $\{P_t\}$ from any initial $P_0$ that results in a constant $G_t = C$. From this line of reasoning it follows that 

\begin{equation}
    \mathbf{Pr}(a_t = 1) = c\mathbf{1}\{ C \geq \phi \} , \text{ for } t>m + 1/\Delta    
\end{equation}
 
 Letting us conclude
 
 \begin{equation}
  \mathbb{E}_{P,V} \big[Q^{\mathbf{R}(\phi)} \big] = c\mathbf{1}\{ C \geq \phi \} + O(1/T)
 \end{equation}
 
 where $\mathbb{E}_{P,V}$ is a short-hand notation for the expectation under the mixture of $\{P_t \}|V=1$ and $\{P_t \}|V=1$ induced by the randomness in $V$.

Furthermore, if  $C < \phi$, then the policy collects no more labels beyond round $m + 1/\Delta$ which implies that $\hat{\mu}_t = \hat{\mu}_{m + 1/\Delta}$ for all $t \geq m + 1/\Delta$. Of course, because no labels are collected after round $m + 1/\Delta$ it is immediate that the long-term expected monitoring risk is at least $\frac{1-\rho}{2}$:

  \begin{equation}
  \label{eqn:}
   \mathbb{E}_{P,V} \big[\ell^{\mathbf{R}(\phi)}(t) | C < \phi  \big] \geq \frac{1-\rho}{2} \text{ if } t \geq m + 2/\Delta
 \end{equation}
 
 Letting us conclude 
 
  \begin{equation}
  \mathbb{E}_{P,V} \big[L^{\mathbf{R}(\phi)} |   C < \phi  \big] \geq  \frac{1-\rho}{2} + O(1/T)
 \end{equation}

We proceed to lower bound the combined loss $\mathcal{L}$ in both the event that $\{C < \phi \}$ and the event that  $\{C \geq \phi \}$

  \begin{equation}
  \mathbb{E}_{P,V} \big[\mathcal{L}^{\mathbf{R}(\phi)} | C \geq \phi \big]  \geq  
  \end{equation}
  \begin{equation}
      \mathbb{E}_{P,V} \big[L^{\mathbf{R}(\phi)}| C \geq \phi  \big] + \mathbb{E}_{P,V} \big[Q^{\mathbf{R}(\phi)}| C \geq \phi  \big] \geq
  \end{equation}
  \begin{equation}
          \mathbb{E}_{P,V} \big[Q^{\mathbf{R}(\phi)}| C \geq \phi  \big] \geq c + O(1/T)
 \end{equation}

  \begin{equation}
  \mathbb{E}_{P,V} \big[\mathcal{L}^{\mathbf{R}(\phi)} | C < \phi \big]  \geq  
      \end{equation}
      \begin{equation}
  \mathbb{E}_{P,V} \big[L^{\mathbf{R}(\phi)}| C < \phi  \big] + \mathbb{E}_{P,V} \big[Q^{\mathbf{R}(\phi)}| C < \phi  \big] \geq
    \end{equation}

  \begin{equation}
  \mathbb{E}_{P,V} \big[L^{\mathbf{R}(\phi)}| C < \phi  \big] \geq \frac{1-\rho}{2} + O(1/T)
 \end{equation}

To complete the proof:

   \begin{equation}
 \sup_{P} \big( \mathbb{E}_{P} \big[\mathcal{L}^{\mathbf{R}(\phi)}  \big] \big) \geq
 \end{equation}
 \begin{equation}
 \mathbb{E}_{P,V} \big[\mathcal{L}^{\mathbf{R}(\phi)}  \big]  \geq  
 \end{equation}
 \begin{equation}
 \mathbf{min} \bigg\{ \mathbb{E}_{P,V} \big[\mathcal{L}^{\mathbf{R}(\phi)}  | C < \phi \big], \text{ }  \mathbb{E}_{P,V} \big[\mathcal{L}^{\mathbf{R}(\phi)} | C \geq \phi  \big] \bigg\} \geq 
 \end{equation}

  \begin{equation}
  \mathbf{min} \bigg\{ c + O(1/T), \text{ } \frac{1-\rho}{2} + O(1/T) \bigg\} \geq 
  \end{equation}
  \begin{equation}
  \mathbf{min} \{c,(1-\rho)/2 \} + O(1/T) 
 \end{equation}

Taking the asymptotic in $T$ yields the result.
 \end{proof}

 Thus, even if the data stream should be easy to monitor because $\Delta$ is small, the RR policy can perform significantly worse than even a naive periodic baseline.
 
We also point out that this result is not dependent on the specific choice of detection window strategy we used (Alg. \ref{alg:detection_windows}). The proof for Thm. 4.1.iii still works even if we change the detection window strategy. 
 
 \begin{remark}
   The proof for Thm. 4.1.iii holds for any \emph{time-bounded} detection window strategy for which there exists some universal $L$ such that $S,S' \subset \{t-L,\dots, t\}$.  
 \end{remark}
 
That we should restrict our detection window strategy to time-bounded windows seems reasonable. We can imagine that looking further and further back in time eventually ceases to be helpful.

\subsubsection{Part (iv)}

We proceed to give a proof for Part (iv) of Theorem \ref{thm:worst_case_regret_main}. This result is heavily based in Le Cam's method \citep{le2012asymptotic}. We begin with Lemma \ref{lemma:MAE_Loss_lecam_bernoulli} which is a Le Cam bound for Bernoulli random variables under MAE loss. This is a standard result which follows directly from the well-established MSE rates. We use Lemma \ref{lemma:MAE_Loss_lecam_bernoulli} in Lemma \ref{lemma:part_iii_of_minimax_theorem_main} which contains the crux of the proof.

\begin{lemma}
\label{lemma:MAE_Loss_lecam_bernoulli}
\label{lemma:bernoulli_lecam}
Let $X^n \sim_{\mathbf{iid} } \mathbf{Bern}(\theta)$ and let the minimization over $\Psi$ take place over the set of all estimators mapping from $\{0,1\}^n$ to $[0,1]$.

$$\inf_{\Psi: \{0,1\}^n  \rightarrow[0,1]} \sup_{\theta } \text{ } \mathbb{E}(|\Psi(X^n) - \theta |)\geq \Theta(1/\sqrt{n})$$
\end{lemma}

\begin{proof}
See references for minimax optimal rates (for example, see \cite{duchi2016lecture}). It is well established that the minimax optimal rate for estimating the mean of a Bernoulli variable under \emph{mean square error}  (MSE) is $\Theta(1/n)$. Elementary modifications to these results yield that under MAE loss the minimax rate is  $\Theta(1/\sqrt{n})$.
\end{proof}

\begin{lemma} 
\label{lemma:part_iii_of_minimax_theorem_main}
No policy can achieve a worst-case expected hinge risk of $\epsilon$ with an average query rate of $\omega(\Delta/\epsilon^3)$.
\end{lemma}

\begin{proof}

For concreteness, we will focus on the hinge loss $R_\mathbf{hinge}$ since the MAE loss will follow immediately from the same proof.

Of course, the following is straightforward for any choice of distribution $F$ over sequences $\{P_t\}$ with support $\mathbf{supp}(F) \subset \mathbf{Lip}(\Delta)$.

\begin{equation}
     \mathbf{max}_{P \in \mathbf{Lip}(\Delta)} \mathbb{E}_{P}[R] \geq \mathbb{E}_{P \sim F} [R]
\end{equation}

The strategy is to construct $F$ for $P_t \in \mathbf{Lip}(\Delta)$ such that monitoring risk $R_\mathbf{hinge}$ requires the same sample complexity as estimating the mean of a Bernoulli under MAE loss. Once this has been done, we will show that 
that any query rate asymptotically lower than order $\Delta/\epsilon^3$ leads to a clear contradiction.



Let $m = 6\epsilon/\Delta$. We proceed to define a generative model for the distribution over $\mathbf{Lip}(\Delta)$. Define distribution $P_Z$ as below:

\[ P_Z(z) = \begin{cases} 
    1/2 & \textbf{if } z = 1    \\
    1/2 & \textbf{if } z = -1\\
      0 & \textbf{else} 
      
   \end{cases}
\]

Sequence $\{ \mu_t\}$ then is generated following:

$$\mu_0 = \frac{1}{2} + 3\epsilon$$

$$\mu_m = \frac{1}{2} + 3\epsilon Z_1$$

$$\mu_{2m} = \frac{1}{2} +  3\epsilon Z_2$$
$$ \vdots $$
$$\mu_{im} = \frac{1}{2} + 3\epsilon  Z_i$$
$$ \vdots $$
$$\mu_{Tm} = \frac{1}{2} + 3\epsilon Z_T$$

The rest of $\{\mu_t\}$ is defined by a linear interpolation between the $\mu_t$ specified above.

It is important to note that for any indices $i,j$ such that $|i-j| > 2m$, that $\mu_i$ is statistically independent of $\mu_j$:

\begin{equation}
\label{eqn:indep_mu}
    \mu_i \perp \mu_j \text{ if } |i-j| > 2m
\end{equation}

For any policy $\pi$, the following lower bounds apply:

\begin{equation}
\label{eqn:lb_worst_1}
    \inf_{\pi} \text{ }  \mathbb{E}[ R_{\mathbf{hinge}}(\mu_t, \hat\mu_t; \rho)] \geq \frac{1}{T} \sum_t \inf_{\pi}  \text{ }  \mathbb{E}[ r_{\mathbf{hinge}}(\mu_t, \hat\mu_t; \rho)] 
\end{equation}

\begin{equation}
\label{eqn:lb_worst_2}
 = \frac{1}{T} \sum_{i=0}^{T/m} \sum_{j=0}^{m} \inf_{\pi} \text{ }  \mathbb{E}[ r_{\mathbf{hinge}}(\mu_{t}, \hat\mu_{t}; \rho)] \text{ for } t = j + iT/m
\end{equation}
\begin{equation}
\begin{split}
\label{eqn:lb_worst_3}
 = \frac{1}{T} \sum_{i=0}^{T/m} \sum_{j=0}^{m} \inf_{\pi} \Big( \mathbf{Pr}(Z_{i-1} = Z_i)\mathbb{E}[ r_{\mathbf{hinge}}(\mu_t, \hat\mu_t; \rho)|Z_{i-1} = Z_i]
 + \mathbf{Pr}(Z_{i-1} \neq Z_i)\mathbb{E}[ r_{\mathbf{hinge}}(\mu_t, \hat\mu_t; \rho)|Z_{i-1} \neq Z_i] \Big)
 \end{split}
\end{equation}

\begin{equation}
\label{eqn:lb_worst_4}
\begin{split}
 = \frac{1}{T} \sum_{i=0}^{T/m} \sum_{j=0}^{m} \inf_{\pi} \Big( \frac{1}{2}\mathbb{E}[ r_{\mathbf{hinge}}(\mu_t, \hat\mu_t; \rho)|Z_{i-1} = Z_i] + 
 \frac{1}{2}\mathbb{E}[ r_{\mathbf{hinge}}(\mu_t, \hat\mu_t; \rho)|Z_{i-1} \neq Z_i] \Big)
 \end{split}
\end{equation}

\begin{equation}
\label{eqn:lb_worst_5}
\begin{split}
     \geq \frac{1}{2T} \sum_{i,j} \inf_{\pi} \Big( \mathbb{E}[ r_{\mathbf{hinge}}(\mu_t, \hat\mu_t; \rho)|Z_{i-1} = Z_i] \Big) + 
      \inf_{\pi}\Big( \mathbb{E}[ r_{\mathbf{hinge}}(\mu_t, \hat\mu_t; \rho)|Z_{i-1} \neq Z_i] \Big)
\end{split}
\end{equation}

\begin{equation}
\label{eqn:lb_worst_6}
 \geq \frac{1}{2T} \sum_{i,j} \inf_{\pi} \Big(\mathbb{E}[ r_{\mathbf{hinge}}(\mu_t, \hat\mu_t; \rho)|Z_{i-1} = Z_i] \Big) 
\end{equation}

\begin{equation}
\label{eqn:lb_worst_7}
 \geq \frac{1}{2T} \sum_{i,j} \inf_{\Psi} \Big( \mathbb{E}[r_{\mathbf{hinge}}(\mu_t, \Psi(\{C_t\}_t)|Z_{i-1} = Z_i]; \rho)]
 \Big) 
\end{equation}

\begin{equation}
\label{eqn:lb_worst_8}
 = \frac{1}{2T} \sum_{i,j} \inf_{\Psi} \Big( \mathbb{E}[r_{\mathbf{hinge}}(\mu_t, \Psi(\{C_\tau\}_{\tau=t-2m}^{t+2m}); \rho)|Z_{i-1} = Z_i]
 \Big) 
\end{equation}

\begin{equation}
\label{eqn:lb_worst_9}
\begin{split}
= \frac{1}{2T} \sum_{i,j}\inf_\Psi \text{ } \mathbb{E}(\Psi(X^{4m}) = \theta) = \\ 
\frac{1}{2}\inf_\Psi \text{ }  \mathbb{E}(\Psi(X^{4m}) = \theta)= \\ \frac{1}{2}\inf_\Psi \text{ }  \mathbf{Pr}(\Psi(X^{4m}) = \theta)
\end{split}
\end{equation}

\begin{equation}
\label{eqn:lb_worst_10}
\geq \Theta(1/\sqrt{m})
\end{equation}
\begin{equation}
\label{eqn:hinge_ub}
= \Theta(\sqrt{\Delta/\epsilon})
\end{equation}

(\ref{eqn:lb_worst_1}) follows from the definition of $r$ and $R$. 

(\ref{eqn:lb_worst_2}) follows from breaking up the sum into a double sum and re-indexing.

(\ref{eqn:lb_worst_3}) follows from the law of total expectation \citep{johnson2000probability}.

(\ref{eqn:lb_worst_4}) follows from the Bernoulli distribution of $Z_i$.

(\ref{eqn:lb_worst_5}) follows from basic properties of optimization \citep{luenberger1984linear}.

(\ref{eqn:lb_worst_6}) follows from the non-negativity of $\ell$.

(\ref{eqn:lb_worst_7}) follows from the fact that the optimal (non-casual) estimator $\Psi$ has access to the entire sequence ${C_t}$ --- in other words all of the labels, even those from the future.

(\ref{eqn:lb_worst_8}) follows from (\ref{eqn:indep_mu}). The labels beyond $2m$ in the future or $2m$ in the past cannot improve the optimal estimator $\Psi$ because they are statistically independent to $\mu_t$ under the generative model for $\{P_t\}$.

(\ref{eqn:lb_worst_9}) follows from the fact that $\mu_t = \rho + 3\epsilon$ with probability $1/2$ and 
$\mu_t = \rho - 3\epsilon$ with probability $1/2$. 

Thus, optimal estimator based on $ \{C_\tau \}_{\tau=t-2m}^{t+2m}$ is no better in expectation than the optimal estimator based on i.i.d. samples from $\mathbf{Bern}(\mu_t)$. This allows us to invoke Lemma \ref{lemma:bernoulli_lecam} to arrive at (\ref{eqn:lb_worst_10}).

(\ref{eqn:hinge_ub}) follows from plugging-in the definition of $m$.

Finally, to complete the proof, recall that by assumption: 
\begin{equation}
\label{eqn:hinge_lb}
\epsilon \geq \mathbb{E}(R_\mathbf{hinge}) \
\end{equation}

Combining (\ref{eqn:hinge_ub}) with (\ref{eqn:hinge_lb}) yields:
\begin{equation}
\Theta\left(\sqrt{\frac{\Delta}{\epsilon}}\right) \leq \epsilon
\end{equation}

Simplify by squaring both sides and multiplying by $\epsilon$. This yields:

\begin{equation}
\Theta(\Delta )\leq \Theta(\epsilon^3)
\end{equation}

From which we conclude that any policy $\pi$ obtaining $\mathbb{E}(Q^{\pi}) = \omega(\Delta/\epsilon^3) = \omega(1)$ actually is querying at a diverging expected rate as $\Delta \rightarrow 0$:
\begin{equation}
    \mathbb{E}(Q^{\pi}) \rightarrow \infty 
\end{equation}

which is of course a contradiction when we know that $\mathbb{E}(Q^{\pi}) \leq 1$.

\end{proof}

\begin{corollary}
\label{corollary:extenstion_from_hinge_to_mae}
No policy can achieve a worst-case expected MAE risk of $\epsilon$ with an average query rate of $\omega(\Delta/\epsilon^3)$
\end{corollary}

\begin{proof}
The proof for Lemma \ref{lemma:part_iii_of_minimax_theorem_main} goes through essentially unchanged for MAE. Simply note that for all $t$:

\begin{equation}
    r_\mathbf{mae}(t) \geq   r_\mathbf{hinge}(t)
\end{equation}

which makes $R_{\mathbf{hinge}}$ a lower bound for $R_{\mathbf{mae}}$.

\end{proof}

\begin{theorem} (Theorem 4.1.iv)

No policy can achieve a worst-case expected loss below $\Omega\left( \Delta^{1/4}\right)$

$$\inf_{(\pi,g) \in \Pi \times \mathcal{G}} \Bigg(\sup_{P \in \mathbf{Lip}(\Delta)} \mathbb{E}_{P}[\mathcal{L}^{\pi}_{g}]\Bigg) = \Omega\left( \Delta^{1/4}\right)$$
\end{theorem}

\begin{proof}

Let $F$ be the distribution over problem instance defined in Lemma \ref{lemma:part_iii_of_minimax_theorem_main}.

We know that:

\begin{equation}
\begin{split}
\inf_{\pi,g } \Bigg(\sup_{P} \mathbb{E}_{P}[\mathcal{L}^{\pi}_{g}]\Bigg)
     \geq 
     \inf_{\pi, g} \mathbb{E}_{P \sim F} [\mathcal{L}_{g}^\pi]
\end{split}
\end{equation}

Define $(\pi^*,g^*)$ as an argmin:

\begin{equation}
  (\pi^*,g^*) \in  \text{arg}\inf_{\pi, g} \text{ }\mathbb{E}_{P \sim F} [\mathcal{L}_{g}^\pi] 
\end{equation}

\begin{equation}\mathbb{E}_{P \sim F} [\mathcal{L}_{g^*}^{\pi^*}] =   \mathbb{E}_{P \sim F} [R_{g^*}^{\pi^*}] + c\cdot \mathbb{E}_{P \sim F} [Q_{g^*}^{\pi^*}] 
\end{equation}

Let $\mathbb{E}_{P \sim F} [R_{g^*}^{\pi^*}] = \epsilon$. Then using Lemma \ref{lemma:part_iii_of_minimax_theorem_main} for hinge loss and its extension, Corollary \ref{corollary:extenstion_from_hinge_to_mae} for MAE:

\begin{equation}\mathbb{E}_{P \sim F} [\mathcal{L}_{g^*}^{\pi^*}] = \epsilon + \Omega(\Delta / \epsilon^3)
\end{equation}

As a final step in the lower bound:

\begin{equation}
    \inf_{\epsilon} \mathbb{E}_{P \sim F} [\mathcal{L}_{g^*}^{\pi^*}]  = \inf_{\epsilon} \big( \epsilon + \Omega(\Delta / \epsilon^3) \big) = \Omega(\Delta^{1/4})
\end{equation}

\end{proof}

\subsection{Proof of Theorem \ref{thm:linear_case_regret_main}}

This result separately analyzes the monitoring risk and the query rates. We introduce the notion of a best-case expected query rate in order to understand the potential upside of the method under favorable conditions. 

\subsubsection{Proof of Theorem \ref{thm:linear_case_regret_main}.i}

\begin{lemma}
Under the conditions of Thm. \ref{thm:worst_case_regret_main}, the following hold  if $(E,\mathsf{d})$ satisfies the linear detection condition with probability $q$:

(i)  If in an estimation problem with MAE loss and
$q < 1 - \epsilon $ then, $$  \frac{3}{13} - \widetilde{O}(\Delta^{1/4}) \leq \inf_{P \in \mathcal{P}} \mathbb{E}_P \left[\frac{\mathcal{Q}^{\mathbf{MLD}}_{g} }{\mathcal{Q}^{\mathbf{PQ}}_{g}}\right]\leq 
\frac{9}{19} -   \widetilde{O}(\Delta^{1/4})$$

\end{lemma}

\begin{proof}

Begin by observing that $Q^{\mathbf{PQ}}$ is constant:

\begin{equation}
    \inf_{P \in \mathcal{P}} \mathbb{E}_P \left[\frac{\mathcal{Q}^{\mathbf{MLD}}_{g} }{\mathcal{Q}^{\mathbf{PQ}}_{g}}\right] = \frac{\inf_P \mathbb{E}_P[\mathcal{Q}^{\mathbf{MLD}}_{g}]}{\mathcal{Q}^{\mathbf{PQ}}_{g}}
\end{equation}

Going forward, we may omit the subscript $g$ for brevity. Recall that $\mathcal{Q}^{\mathbf{PQ}} = \frac{n}{(\alpha+1)n}$. Furthermore, recall from (\ref{eqn:Q_mld_formula_full}) that: $\inf_P \mathbb{E}_P\mathcal{Q}^{\mathbf{MLD}} = \frac{n}{(\alpha+1)n + \mathfrak{n}_{\max}}$.

The ratio comes out to be:

\begin{equation}
    \frac{(\alpha+1)n}{(\alpha +1)n + \mathfrak{n}_{\max} } = \frac{1}{1 + \zeta}
\end{equation}

where $\zeta$ is defined as:
\begin{equation}
    \zeta = \frac{\mathfrak{n}_{\max}}{(\alpha+1)n}
\end{equation}

Thus, the key in understanding the ratio lies in understanding $\zeta$. To proceed, we will plug-in the values of $\alpha$, $n$ and $\mathfrak{n}_{\max}$ in terms of $\epsilon$ and $\Delta$ in order to simplify the expression.

Recall:

\begin{equation}
    \alpha = \frac{\epsilon^3}{15\Delta\log(2\rho/\epsilon)}
\end{equation}

\begin{equation}
    n = \frac{9\log(2\rho/\epsilon)}{2\epsilon^2}
\end{equation}

where the above two equations follow by construction (refer to A.1) and
\begin{equation}
    \mathfrak{n}_{\max} =  \mathsf{b}\epsilon/\Delta
\end{equation}
for some $1/3 \leq \mathsf{b} \leq 1$. This follows from Lemma \ref{lemma:bounds_for_query_period_mld}.

Plugging these into $\zeta$ and simplifying yields a constant term and a term vanishing in $\Delta$:

\begin{equation}
    \zeta = \frac{30\mathsf{b}}{9} + \widetilde{O}(\Delta^{1/4})
\end{equation}

Now, plugging $\zeta$ back into the ratio gives us:

\begin{equation}
   \inf_{P \in \mathcal{P}} \mathbb{E}_P \left[\frac{\mathcal{Q}^{\mathbf{MLD}} }{\mathcal{Q}^{\mathbf{PQ}}}\right] = \frac{9}{30\mathsf{b} + 9} - \widetilde{O}(\Delta^{1/4})
\end{equation}

Applying the upper and lower bounds on $\mathsf{b}$ completes the proof.
\end{proof}

\subsubsection{Proof of Theorem \ref{thm:linear_case_regret_main}.ii}

\begin{theorem} 
Under the conditions of Thm. \ref{thm:worst_case_regret_main}, the following hold  if $(E,\mathsf{d})$ satisfies the linear detection condition with probability $q$:

(ii) if using hinge risk and $q < 1 - \epsilon $ then,

$$ \inf_{P \in \mathcal{P}} \mathbb{E}_P \left[\frac{\mathcal{Q}^{\mathbf{MLD}}_{g} }{\mathcal{Q}^{\mathbf{PQ}}_{g}}\right] =  \widetilde{O}(\Delta^{1/4})$$

\end{theorem}

\begin{proof}
This is a trivial corollary to Lemma \ref{lemma:avg_q_decision}. 

For any choice of generative model for $P$ (such as $\mathcal{S}$ from Lemma \ref{lemma:avg_q_decision}):

\begin{equation}
    \inf_{P}\left( \mathbb{E}_P[\mathcal{Q}^{\mathbf{MLD}}]\right) \leq \mathbb{E}_{P\sim \mathcal{S}}[\mathcal{Q}^{\mathbf{MLD}}]
\end{equation}
\end{proof}

\subsubsection{Proof of Theorem \ref{thm:linear_case_regret_main}.iii}

\begin{theorem} 
Under the conditions of Thm. \ref{thm:worst_case_regret_main}, the following hold  if $(E,\mathsf{d})$ satisfies the linear detection condition with probability $q$:

(iii) if  $q \geq 1 - \epsilon$, then there exist problem instances $P$ for which $Q_g^{\mathbf{MLD}} = 0$ almost surely: $ \frac{\mathcal{Q}^{\mathbf{PQ}}_{g} }{\inf_{P \in \mathcal{P}} \mathbb{E}_{P}[\mathcal{Q}^{\mathbf{MLD}}_{g}]} = \infty $
\end{theorem}

\begin{proof}
Consider a problem instance in which the linear detection condition holds with $\mathcal{N} = 0$
and $\mu_t = 1$ for all $t$. In this case, the standard error of the forecast for \textsc{MLDemon}'s linear model is 0: 

\begin{equation}
    \mathsf{se} = 0
\end{equation}

Based on this, we obtain perfect confidence surrounding our estimate: $p_\text{det} = 1$

Therefore, $p_{t} \geq 1 - \epsilon$ for all $t$. The conclusion from this is that \textsc{MLDemon} never queries for another batch of labels.

\end{proof}

\subsection{Proofs of Lemma \ref{lemma:avg_q_decision} and Theorem \ref{thm:avg_case_main}}

\subsubsection{Proof of Lemma \ref{lemma:avg_q_decision}}
We proceed to finish our mathematical details with proving the average-case analysis. Refer to the main text for the definition of random walk that generates model $\mathcal{S}$ for the drift. We begin with  Lemma \ref{lemma:avg_q_decision}, since Theorem \ref{thm:avg_case_main} is a simple corollary thereof.

\begin{lemma} (Lemma \ref{lemma:avg_q_decision})

  For decision problems with hinge risk under model $\mathcal{S}$, \textsc{MLDemon} achieves an expected monitoring hinge risk $O(\epsilon)$ with an amortized query rate  $\widetilde{O}\left(\Delta/\epsilon^2\right)$. 
\end{lemma}
\begin{proof}

 The bound on monitoring risk follows from Lemma \ref{lemma:worst_case_tradeoff}. More concretely, the monitoring risk is no greater than $2\epsilon$ (see Remark \ref{remark:mld_risk_2_eps}). We shall also make use of the decision margin $\ell_t$ (\ref{def:dec_margin}) in this proof. We proceed compute the amortized query rate.

Notice that as $t \rightarrow \infty$ we have that $\mu_t \rightarrow \mathbf{Unif}(0,1)$ in distribution. The policy's estimate $\hat{\mu}_t$ takes on values in $\{0, 1/n, 2/n, ..., 1\}$. For any fraction $\mathfrak{u} \in \{0, 1/n, 2/n, ..., 1\}$ we can lower bound the \emph{steady-state} probability that $\hat \mu_t$ takes on $\mathfrak{u}$:

\begin{equation}
\label{eqn:q_prob_mu_hat}
\lim_{t \rightarrow \infty} \mathbf{Pr}[\hat{\mu}_t = \mathfrak{u}] = \Theta(1/n)
\end{equation}

We shall use the short-hand notation $\mathbf{Pr}_\infty[\cdot] = \lim_{t \rightarrow \infty} \mathbf{Pr}[\cdot]$.

Furthermore, note that for any $\mathfrak{u} \leq 1/2$, we still have

\begin{equation}
\label{eqn:q_prob_mu_hat_w_rho}
\mathbf{Pr}_\infty[|\hat{\mu}_t-\rho| = \mathfrak{u}] = \Theta(1/n) \text{ for } \mathfrak{u} \leq 1/2
\end{equation}

Since for any choice of $\rho$ (and  $\mathfrak{u} \leq 1/2$):

\begin{equation}
\label{eqn:q_prob_mu_hat_2_sided_ineq}
 \mathbf{Pr}_\infty[\hat{\mu}_t = \mathfrak{u}] \leq  \mathbf{Pr}_\infty[|\hat{\mu}_t-\rho| = \mathfrak{u}] \leq 2 \mathbf{Pr}_\infty[\hat{\mu}_t = \mathfrak{u}]
\end{equation}


Based on \ref{eqn:q_prob_mu_hat_w_rho}, we can see steady-state expectation of $\ell_t$:

\begin{equation}
   \lim_{t \rightarrow \infty} \mathbb{E}[\ell_t] \geq \sum_{i=\mathbf{ceil}(\epsilon)}^{n/2} \frac{i}{n} \left(\mathbf{Pr}\left[\hat{\mu}_t =\frac{i}{n}\right] \right) 
\end{equation}

\begin{equation}
   = \sum_{i=\mathbf{ceil}(\epsilon)}^{n/2} i \cdot \Theta \left({\frac{1}{ n^2}} \right) = \Theta\left( 1\right)
\end{equation}

Recall the satisfiability condition \ref{eqn:main_hoeff_ineq}:

\begin{equation}
  \nu_t + \delta \leq \ell_t + 2\epsilon
\end{equation}

We must ask ourselves, how large does $\nu_t$ become now? Recall that $\nu_t = \Delta(\tau + (n+1)/2)$. In order for MLD to commence a query batch, it must be the case that:

\begin{equation}
    \nu_t \sim \ell_t = \Theta(1)
\end{equation}

Implying

\begin{equation}
    \Delta\tau  = \Theta(1)
\end{equation}

Starting with the rate $\tau = \Theta(1/\Delta)$ and then following an argument nearly identical to that of (\ref{eqn:start_of_Q_analysis})-(\ref{eqn:end_of_Q_analysis}), we arrive at:

\begin{equation}
    \mathbb{E}_{P \sim \mathcal{S}}[Q^{\mathbf{MLD}}] =\widetilde{\Theta}(\Delta/\epsilon^2) = \widetilde{\Theta}(\Delta^{1/2})
\end{equation}




\end{proof}

\subsubsection{Proof of Theorem \ref{thm:avg_case_main}}





\begin{theorem}
Let $\mathcal{S}$ be the distribution over problem instances implied by the stochastic model. For any model $f$ and any detector $g$, on the decision problem with hinge risk:

 $$ \frac{\mathbb{E}_{\mathcal{S}} \mathcal{L}_g^{\mathbf{MLD}}}{ \mathbb{E}_{\mathcal{S}}  \mathcal{L}_g^{\mathbf{PQ}}} = \widetilde{O}(\Delta^{1/12})$$
 
\end{theorem}

\begin{proof}
For both $\mathbf{MLD}$ and $\mathbf{PQ}$, the choice of $g$ affects both $R$ and $Q$ up to constant factors. 

Following Lem. \ref{lemma:PQ_final_step}, but replacing the worst-case expected amortized query of $\widetilde{O}(\Delta / \epsilon^3)$ with the expectation under $S$ of $\widetilde{O}(\Delta / \epsilon^2)$ yields a combined loss:

\begin{equation}
    \mathbb{E}_\mathcal{S}[\mathcal{L}^\mathbf{MLD}] = \widetilde{O} (\Delta^{1/3})
\end{equation}

On the other hand, we know that $PQ$ is not data dependent, so the expected loss on $\mathcal{S}$ is the same as the worst-case. From Lem. \ref{lemma:worst_case_tradeoff}:

\begin{equation}
    \mathbb{E}_\mathcal{S}[\mathcal{L}^\mathbf{PQ}] =  \sup_{P\in \mathbf{Lip}(\Delta)}\mathbb{E}_P[\mathcal{L}^\mathbf{PQ}] = \widetilde{\Theta} (\Delta^{1/4})
\end{equation}

From which we obtain the rate improvement ratio $\widetilde{O}(\Delta^{1/12})$.

\end{proof}

\section{VISUALIZATIONS}

 Here, we provide some additional results with the goal of helping visualize the algorithms and give some extra intuition to the behaviors of the algorithms. These visualizations capture sequential behavior for a single random seed. Recall that the batch size is held constant in all experiments (at $n = 35$). 
 
  In Figs. 3 \& 4 below, the orange signal is the true $\mu_t$ over time. The blue spikes indicate that the policy elected to query a batch of labels. The blue signals in 3c and 4e are the detection signal $g_t$. 
  
 Fig. 3 illustrates the unintutitve yet possible scenario in which RR performs worse than both PQ and \textsc{MLDemon} even though the  detection signal is correlated and informative (3c).  RR (3a) simply waits too long to query because the detection signal's largest spikes are near the end of the stream. The result is that RR performs worse than PQ and \textsc{MLDemon}. Such drifts showcase the need for adaptivity in how the policy interprets the detection signal.
 
 \begin{figure}[H]
\centering
\label{fig:3}

\begin{subfigure}{.33\textwidth}
    \centering
    \includegraphics[width=1\textwidth]{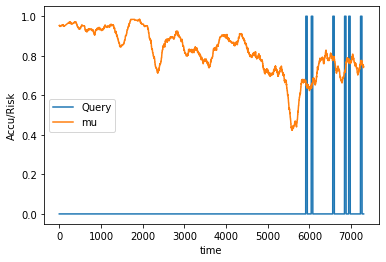}
    \caption{\scriptsize  RR at $\phi = 0.2$. RR queries for $6$ batches \\ at empirical risk $R = 886.6$.  }
\end{subfigure}%
\begin{subfigure}{.33\textwidth}
    \centering
    \includegraphics[width=1\textwidth]{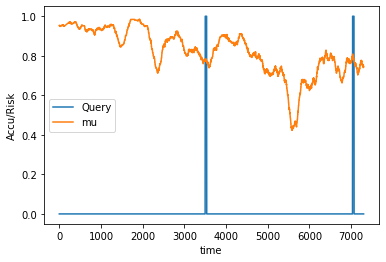}\caption{  \scriptsize \textsc{MLDemon} \& PQ (same behavior).  Both query for $2$ batches at empirical risk $R = 516.2$.}
\end{subfigure}%
\begin{subfigure}{.33\textwidth}
    \centering
    \includegraphics[width=\textwidth]{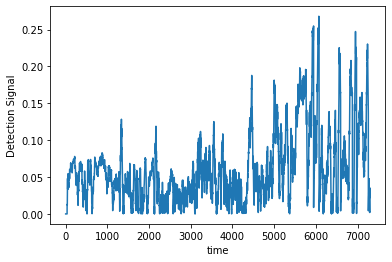}\caption{ Detection signal $G_t$ \\ }
\end{subfigure}
\caption{\footnotesize Sequential behavior on a single random seed for \texttt{SPAM-CORPUS}}
\end{figure}

 \twocolumn

 \begin{figure}[H]
 \label{fig:4}
\centering
\begin{subfigure}{.33\textwidth}
    \centering
    \includegraphics[width=1\textwidth]{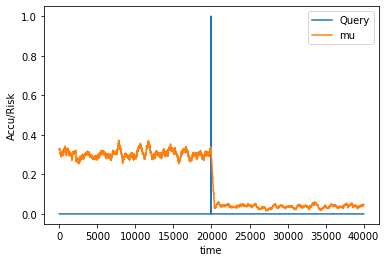}
    \caption{  RR at $\phi = 5$. }
\end{subfigure}%

\begin{subfigure}{.33\textwidth}
    \centering
    \includegraphics[width=1\textwidth]{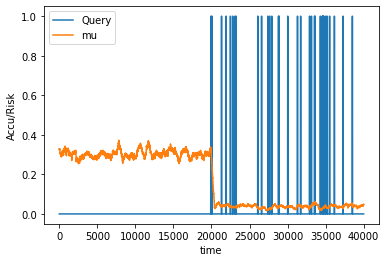}\caption{   RR at $\phi = 3$.}
\end{subfigure}%

\begin{subfigure}{.33\textwidth}
    \centering
    \includegraphics[width=1\textwidth]{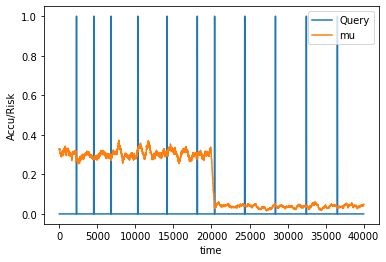}\caption{ MLD at $\alpha=64$.}
\end{subfigure}

\begin{subfigure}{.33\textwidth}
    \centering
    \includegraphics[width=1\textwidth]{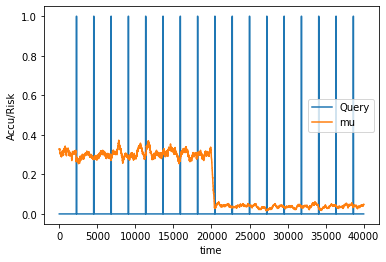}\caption{ PQ at $\alpha=64$.}
\end{subfigure}

\begin{subfigure}{.33\textwidth}
    \centering
    \includegraphics[width=1\textwidth]{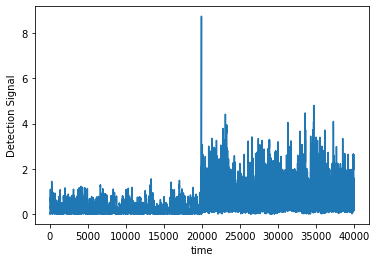}
    \caption{Detection signal $G_t$ \\}
\end{subfigure}

\caption{\footnotesize Sequential behavior on a single random seed for \texttt{FACE-RECOG}}
\end{figure}

 Interestingly, we can see that this example also illustrates an instance in which \textsc{MLDemon} and PQ behave the same. \textsc{MLDemon} needs at least a few samples in order to gain enough confidence in the model fit to extend the period. In the limit of few queries, \textsc{MLDemon} always performs like PQ.

 In Fig. 4, all 3 policies attain roughly the same empirical risk. This is fairly intuitive from seeing when the query batches take place. However, we can see that the different policies use vastly different numbers of query batches to get here.

 Compare 4a and 4b to see how RR picks the optimal time to query when only using 1 batch, but neglects the beginning of the stream as $\phi$ increases (resulting in label waste). This happens because the detection signal (4e) correctly detects the large drop in accuracy, but has a higher baseline after the drop than before.

  Furthermore, observe how PQ uses 17 query batches in 4d while \textsc{MLDemon} uses only 11 in 4c. Furthermore, if we look closely at 4c, we can observe the way that \textsc{MLDemon} \emph{extends} the query period after the first few batches, but reverts back when the detection signal spikes. This visualization neatly showcases \textsc{MLDemon}'s behavior and how \textsc{MLDemon} builds an advantage over PQ.

\end{document}